\pdfoutput=1
\PassOptionsToPackage{table}{xcolor}
\documentclass[]{opendatalab}

\usepackage[T1]{fontenc}
\usepackage[utf8]{inputenc}
\usepackage{tabularx}
\usepackage{float}
\usepackage{amssymb}

\title{RxnExBench: Benchmarking Vision-Language Models for Structured Chemical Reaction Extraction from Scientific Literature}

\author[1,*]{Jiaxing Sun}
\author[1,*]{Junyuan Gao}
\author[2,*]{Ruikang Zhang}
\author[3,*]{Hao Zheng}
\author[4,1,5,*]{Lingli Ge}
\author[6,*]{Xiao An}
\author[1,*,\dagger]{Jiang Wu}
\author[7,5]{Huazheng Zeng}
\author[4,1]{Jiahe Song}
\author[8]{Chen Zhu}
\author[1]{Haote Yang}
\author[9]{Jingchao Wang}
\author[1]{Lu Chen}
\author[1]{Yuanyuan Cao}
\author[1]{Ruijie Zhang}
\author[1]{Shengxin Lu}
\author[10]{Weijia Li}
\author[1]{Lijun Wu}
\author[1,\ddagger]{Conghui He}

\affiliation[1]{Shanghai Artificial Intelligence Laboratory}
\affiliation[2]{Peking University}
\affiliation[3]{South China Normal University}
\affiliation[4]{Shanghai Jiao Tong University}
\affiliation[5]{Shanghai Innovation Institute}
\affiliation[6]{Wuhan University}
\affiliation[7]{Fudan University}
\affiliation[8]{East China University of Science and Technology}
\affiliation[9]{East China Normal University}
\affiliation[10]{Tsinghua Shenzhen International Graduate School}

\contribution[*]{Equal contribution.}
\contribution[\dagger]{Project lead.}
\contribution[\ddagger]{Corresponding author.}

\abstract{\textit{Screen Tables} and \textit{Substrate Scopes} are information-dense figures widely used in synthetic chemistry literature, yet automated extraction of structured reaction data from them remains challenging. We introduce \textbf{RxnExBench}, a multimodal benchmark for fine-grained reaction-level evaluation. Built from high-impact chemistry journals, it contains 3,565 expert-annotated reactions and 2,361 molecular structures. The dataset adopts a unified schema with 15 reaction roles and 26 core fields, supporting downstream AI-for-Chemistry tasks such as retrosynthetic planning, yield prediction, and reaction condition recommendation. RxnExBench further provides a reasoning-driven difficulty taxonomy, a modular annotation pipeline, and the \textsf{Oracle-MolRec} protocol, which disentangles molecular recognition from higher-level reasoning. Evaluation of 9 state-of-the-art VLMs shows that performance degrades substantially as schema complexity and reasoning demands increase. Under the \textsf{Oracle-MolRec} setting, results further reveal that accurate molecular recognition alone is insufficient, with multimodal integration and chemistry-aware reasoning remaining the primary bottlenecks. We release all data and code to facilitate future research in chemistry-aware multimodal understanding and AI-for-Chemistry dataset construction. The dataset and code will be released at \url{https://github.com/opendatalab/RxnExBench}.
}

\date{\today}
\metadata[Equal Contribution ($*$)]{Jiaxing Sun, Junyuan Gao, Ruikang Zhang, Hao Zheng, Lingli Ge, Xiao An, Jiang Wu }
\metadata[Project Lead ($\dagger$)]{Jiang Wu, \email{wujiang@pjlab.org.cn}}
\correspondence{Conghui He, \email{heconghui@pjlab.org.cn}}

\begin{document}

\maketitle
\fancyhead[L]{\nouppercase{\textbf{RxnExBench}}}

\section{Introduction}
\label{sec:intro}
\begin{figure}[!t]
    \centering
    \includegraphics[width=0.8\linewidth]{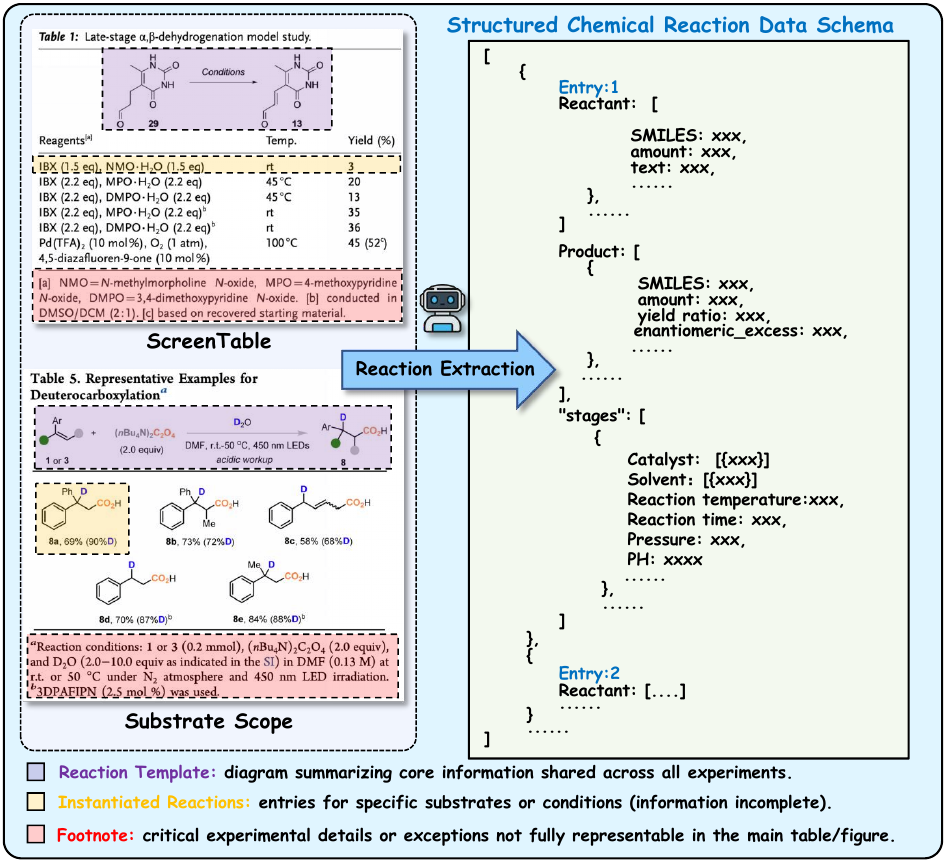}
    \caption{\textit{Screen Table} and \textit{Substrate Scope} are information-rich figure types commonly used in chemical papers to present chemical reactions. A typical \textit{Screen Table} or \textit{Substrate Scope} usually contains: a ``reaction template'' presented as an image (purple box), ``instantiated'' chemical reactions in a tabular or substrate array format (yellow box), and a footnote for important supplementary information that cannot be fully displayed in the figure (red box). The task of \textbf{RxnExBench} is to extract structured information for all chemical reactions from these figures according to the defined schema.}
    \label{fig:fig1_REB_example}
\end{figure}

\begin{table*}[t]
\centering
\small
\setlength{\tabcolsep}{5pt}
\caption{Comparison of related benchmarks and our \textbf{RxnExBench}. \textsf{StructRxnEx} indicates that the benchmark's task is structured chemical reaction information extraction; \textsf{R-group Resolved} means that for all instantiated chemical reactions, the R-groups in all molecular structure diagrams have been resolved; ``\textsf{CLS} figures" means that the benchmark contains the ``complex-line-substrate" figures (see Figure~\ref{fig:fig1_REB_multi_rxn} for details).}
\resizebox{\linewidth}{!}{
\begin{tabular}{lccccccccc}
\toprule
\textbf{\begin{tabular}[c]{@{}c@{}}Benchmark\end{tabular}} &
\textbf{\begin{tabular}[c]{@{}c@{}}Figure Types\end{tabular}} &
\textbf{\begin{tabular}[c]{@{}c@{}}StructRxnEx\end{tabular}} &
\textbf{\begin{tabular}[c]{@{}c@{}}R-group\\ Resolved\end{tabular}} &
\textbf{\begin{tabular}[c]{@{}c@{}}\textsf{CLS} Figures\end{tabular}} &
\textbf{\begin{tabular}[c]{@{}c@{}}Difficulty\\Levels\end{tabular}} &
\textbf{\begin{tabular}[c]{@{}c@{}}\# Schema\\Fields
\end{tabular}} &
\textbf{\begin{tabular}[c]{@{}c@{}}\#Rxn\\\end{tabular}} &
\textbf{\begin{tabular}[c]{@{}c@{}}Avg Tokens\\/Fig\end{tabular}} &
\textbf{\begin{tabular}[c]{@{}c@{}}Avg Rxn\\/Fig\end{tabular}} \\
\midrule
ChemTable &
Screen Table, Substrate Scope &
$\times$ & $\times$ & -- & -- & -- & -- & -- & -- \\

MERMES-T24 &
Substrate Scope &
$\times$ & $\times$ & -- & -- & -- & -- & -- & -- \\

ChemEAGLE &
Substrate Scope &
$\checkmark$ & $\checkmark$ & $\times$ &
Level 3 & 11 & 1120 &
1{,}179.91 & 9.21 \\

MerMaid &
Screen Table &
$\times$ & $\checkmark$ & $\times$ &
Level 1,2 & 15 & 1343 &
2{,}707.64 & 12.32 \\
\midrule
\textbf{RxnExBench} &
\textbf{Screen Table, Substrate Scope} &
\textbf{$\checkmark$} & \textbf{$\checkmark$} & \textbf{$\checkmark$} &
\textbf{Level 1,2,3} & \textbf{26} & \textbf{3565} &
\textbf{7{,}205.28} & \textbf{13.40} \\
\bottomrule
\end{tabular}
}
\label{tab:Comparison_bench}
\end{table*}

Recent years have witnessed significant progress in applying artificial intelligence to tasks such as chemical reaction prediction~\cite{joung2025electron}, reaction condition optimization~\cite{wang2024identifying}, and retrosynthetic planning~\cite{xiong2025bridging,deng2025rsgpt,xu2025unified}. However, a critical bottleneck in current AI-for-Chemistry research remains the scarcity of large-scale, high-quality chemical reaction datasets~\cite{ding2025survey,zhong2024recent}. While the vast chemical literature contains abundant reaction information, it is predominantly presented in unstructured formats, such as complex figures and tables, making it unsuitable for direct use in training machine learning models. Consequently, automatically extracting structured reaction data from scientific papers has become an important and actively pursued research direction.

Existing efforts have largely focused on textual reaction information extraction~\cite{guo2021automated,zhong2023reaction}, molecular structure recognition~\cite{qian2023molscribe,wang2025gtr}, and reaction scheme parsing~\cite{qian2023rxnscribe,chen2025towards,song2025rxncaption}. Yet, these highly informative but complex graphical formats—specifically \textit{Screen Tables} and \textit{Substrate Scopes}—remain largely underexplored, despite being ubiquitous in synthetic chemistry papers.

As illustrated in Figure \ref{fig:fig1_REB_example}, \textit{Screen Tables} compare closely related reactions under varying conditions to identify optimal parameters, while \textit{Substrate Scopes} present reaction outcomes across diverse reactants under fixed conditions to demonstrate reaction generality. Both formats encode dense experimental knowledge through compact notations, domain-specific abbreviations, and embedded molecular diagrams. Consequently, they represent highly valuable yet notoriously challenging sources for automated extraction. Accurate parsing therefore requires not only visual understanding and OCR capabilities, but also chemistry-aware symbolic interpretation and cross-modal reasoning.

The recent rise of Vision-Language Models (VLMs), such as Gemini~\cite{team2023gemini}, Qwen-VL~\cite{bai2025qwen2}, and InternVL~\cite{chen2024internvl}, has shown great promise in scientific document understanding, with notable advances in layout analysis~\cite{wei2024general,niu2025mineru2}, molecular graph recognition~\cite{wang2025gtr}, reaction diagram parsing~\cite{song2025rxncaption}, and property extraction~\cite{yan2025biominer}. Nevertheless, the performance of VLMs in comprehending and structuring reactions from \textit{Screen Tables} and \textit{Substrate Scopes} has not been systematically evaluated. As shown in Table~\ref{tab:Comparison_bench}, existing benchmarks exhibit clear limitations: MERMES~\cite{leong2024automated} focuses solely on condition extraction without modeling full reactions; ChemTable~\cite{zhou2025benchmarking} addresses table structure but ignores reaction semantics; while ChemEAGLE~\cite{chen2025multi} and MerMaid~\cite{leong2025mermaid} suffer from narrow chart coverage and oversimplified evaluation protocols, ultimately failing to reflect the true complexity of real-world chemical literature.

\begin{figure*}[!t]
    \centering
    \includegraphics[width=1\linewidth]{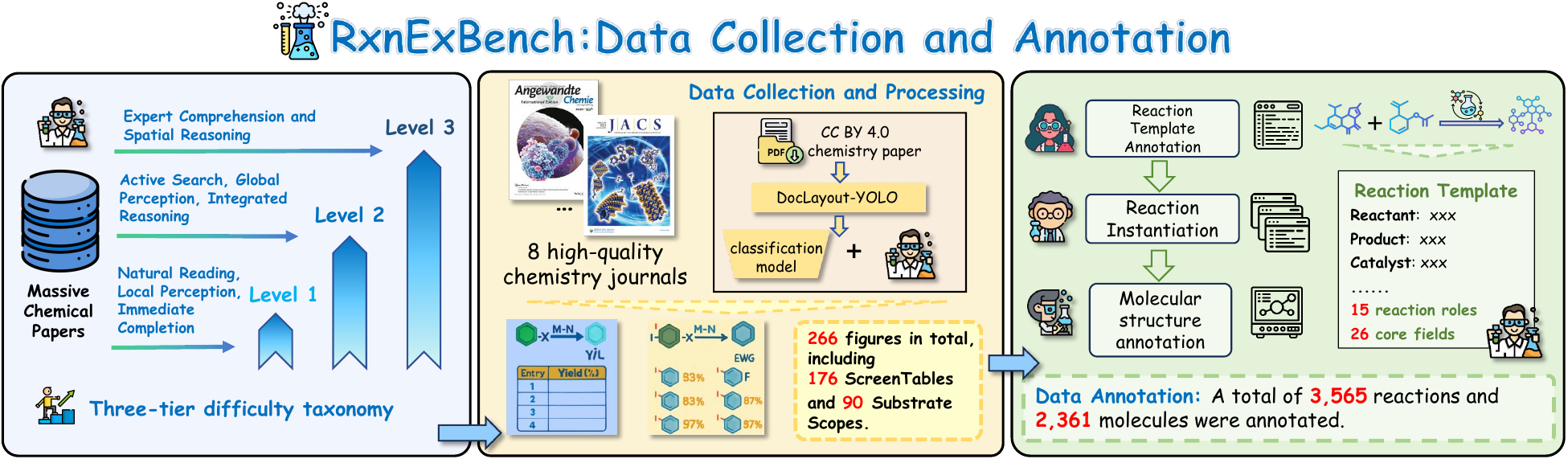}
    \caption{Data collection and annotation of RxnExBench}
    \label{fig:fig1_Data_Collection_and_Annotation}
\end{figure*}

To address this critical gap, we introduce \textbf{RxnExBench}, a comprehensive benchmark designed to evaluate VLMs on structured reaction extraction from \textit{Screen Tables} and \textit{Substrate Scopes}. Shifting the paradigm from figure-level evaluation to fine-grained, reaction-level understanding, RxnExBench adopts a unified schema featuring 15 reaction roles and 26 core fields, explicitly designed to support downstream AI-for-Chemistry tasks. Furthermore, we categorize the extraction process into three reasoning-driven difficulty levels, providing a granular diagnosis of model capabilities. Compared to existing benchmarks, RxnExBench introduces substantially greater complexity, multimodal integration, and reasoning diversity. Our main contributions are summarized as follows:

\begin{enumerate}
\item \textbf{A systematic framework for chemistry-aware extraction and reasoning evaluation.} To tackle the significant challenge of extracting structured chemical reactions from \textit{Screen Tables} and \textit{Substrate Scopes}, we formalize this end-to-end task and design a comprehensive 26-field schema aligned with major AI-for-Chemistry applications. Recognizing that this extraction demands holistic multimodal reasoning beyond basic chart parsing, we further introduce a three-tier classification (local reading, global integration, and R-group-aware spatial reasoning) grounded in chemistry, which empirically correlates with VLM error rates.

\item \textbf{A modular high-quality annotation pipeline.} To support the complex demands of this task, we develop a multi-stage labeling workflow (template annotation, instantiation, Ketcher-based molecule drawing, RDKit canonicalization). Executed by 40 chemistry-trained annotators with three rounds of cross-validation, this pipeline yields a robust dataset of 3,565 reactions, 2,361 molecular structures and 53,659 annotated fields.

\item \textbf{Empirical study and the \textsf{Oracle-MolRec} protocol.} We benchmark 9 state-of-the-art VLMs and design the \textsf{Oracle-MolRec} inference protocol to explicitly decouple fundamental molecular recognition from higher-level reasoning. Our evaluation shows that, while VLMs handle basic chart parsing, they struggle with holistic chemistry-grounded integration as schema demands grow; moreover, \textsf{Oracle-MolRec} reveals that even with perfect molecular recognition, higher-level reasoning remains the dominant bottleneck for fully automated extraction.
\end{enumerate}

\section{Related Work}
\label{sec:related_works}

Large-scale structured chemical reaction datasets extracted from the scientific literature~\cite{schneider2016big,schneider2016s} have significantly advanced AI-for-Chemistry research. Recent efforts in automated extraction have predominantly focused on \textit{text-based} approaches, including deep learning methods~\cite{guo2021automated} and, more recently, large language model (LLM)-driven techniques~\cite{zhong2023reaction}. However, a substantial portion of critical reaction information in chemistry papers is conveyed through \textit{images}, motivating growing interest in \textit{image-based} reaction extraction. Several works have targeted general reaction diagram parsing~\cite{qian2023rxnscribe,chen2025towards,song2025rxncaption}, yet \textit{Screen Tables} and \textit{Substrate Scopes}, which are two highly informative and challenging chart types, remain underexplored despite their ubiquity in synthetic chemistry literature.

As summarized in Table~\ref{tab:Comparison_bench}, existing datasets exhibit notable limitations.
\textbf{MERMES}~\cite{leong2024automated} constructs the MERMES-T24 dataset from 65 papers and evaluates five VLMs on reaction diagram parsing and cross-modal reasoning. However, its annotations focus only on reaction conditions and a single substrate, making it suitable for partial element extraction but inadequate for assessing complete reaction reconstruction.
\textbf{MerMaid}~\cite{leong2025mermaid} covers broader reaction scenarios and introduces SMILES fields to evaluate molecular-to-SMILES conversion. It specifically targets R-group-dependent reasoning in \textit{Screen Tables}, but does not address \textit{Substrate Scopes} or multi-condition comparison.
\textbf{ChemEAGLE}~\cite{chen2025multi} evaluates VLMs on \textit{Substrate Scope} understanding, yet considers only a limited set of reaction roles, falling short in assessing holistic, multi-modal integration into full reaction records.
\textbf{ChemTable}~\cite{zhou2025benchmarking} includes both \textit{Substrate Scopes} and \textit{Screen Tables}, but primarily measures table-structure comprehension rather than structured chemical reaction extraction.
\begin{table}[t]
\centering
\small
\caption{Statistics of reaction figures and counts.}
\resizebox{0.8\linewidth}{!}{
\setlength{\tabcolsep}{6pt}
\begin{tabular}{lccccc}
\toprule
\textbf{Type} & \textbf{\#Figures} & \textbf{\#Min Rxn} & \textbf{\#Max Rxn} & \textbf{\#Avg Rxn} & \textbf{\#Total Rxn} \\
\midrule
Screen Table      & 176 & 3 & 29 & 11.53 & 2,029 \\
Substrate Scope   & 90  & 3 & 44 & 17.07 & 1,536 \\
\midrule
Level 1   & 60  & 4 & 19 & 9.85 & 591 \\
Level 2      & 90 & 3 & 29 & 12.12 & 1,091 \\
Level 3   & 116  & 3 & 44 & 16.23 & 1,883 \\
\midrule
\textbf{Total}   & \textbf{266} & \textbf{3} & \textbf{44} & \textbf{13.40} & \textbf{3,565} \\
\bottomrule
\end{tabular}
}

\label{tab:reaction-stats}
\end{table}

\section{RxnExBench}
\label{sec:benchmark}

\begin{figure}[!t]
    \centering
    \small
    \includegraphics[width=0.8\linewidth]{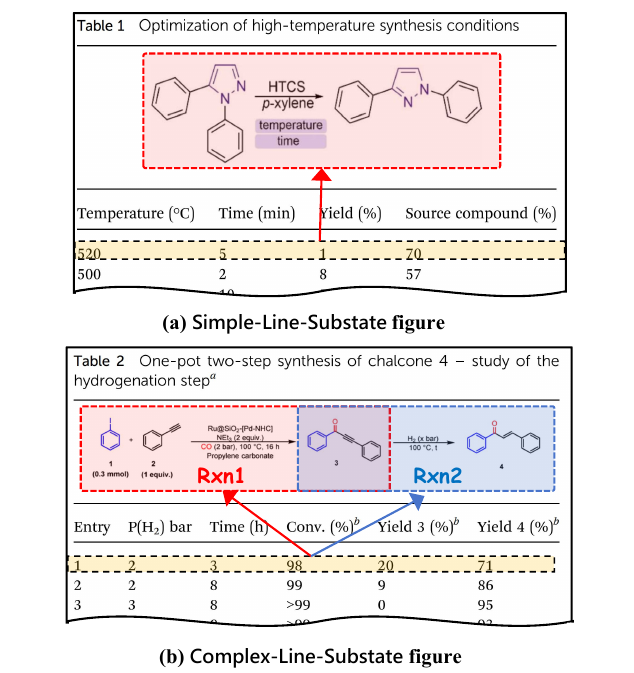}
    \caption{In \textit{Screen Tables}, each row typically corresponds to one instantiated reaction; in \textit{Substrate Scopes}, each substrate likewise. Figures where all rows/substrates meet this condition are called ``simple-line-substrate'' (\textsf{SLS}) figures, as shown in (a). However, some rows or substrates may contain multiple reactions; these are termed ``complex-line-substrate'' (\textsf{CLS}) figures, as shown in (b).}
    \label{fig:fig1_REB_multi_rxn}
    \vspace{-1.0\baselineskip}
\end{figure}

We present \textbf{RxnExBench}, a multimodal benchmark designed to comprehensively and rigorously evaluate Vision-Language Models (VLMs) on their ability to understand, interpret, and extract \textit{structured chemical reaction information} from two critical yet understudied chart types in synthetic chemistry literature: \textit{Screen Tables} and \textit{Substrate Scopes}.

\begin{figure*}[!t]
    \centering
    \includegraphics[width=1\linewidth]{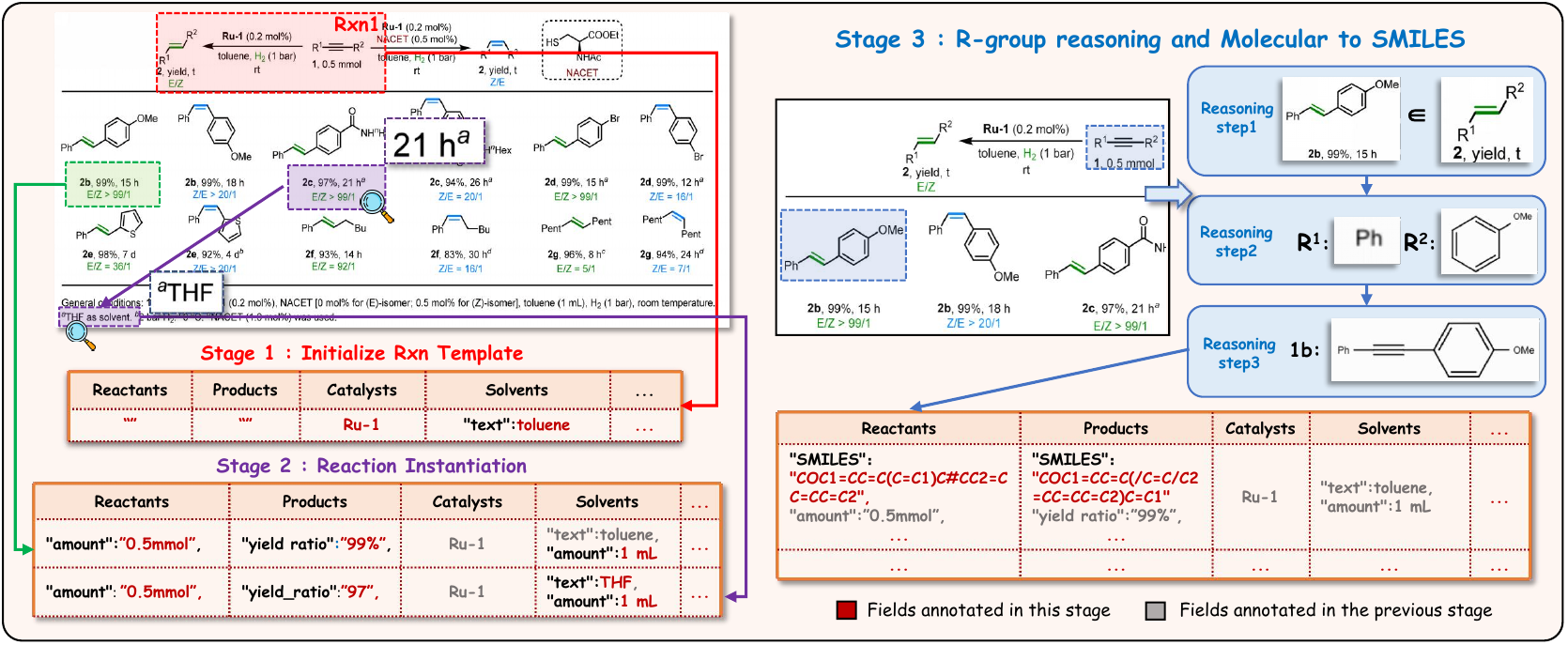}
    \caption{Three stages of structured chemical reaction extraction from \textit{Screen Tables} and \textit{Substrate Scopes}, describing the processes of human expert reading, VLM understanding, and manual data annotation.}
    \label{fig:fig1_REB_reasoning_stages}
    \vspace{-1.0\baselineskip}
\end{figure*}

\subsection{Structured Reaction Schema}
\label{subsec:schema}

Inspired by prior works~\cite{zhong-etal-2023-reaction,cas2025scifinder}, we define a structured schema for representing chemical reactions (Figure~\ref{fig:fig1_REB_example}). Each reaction consists of three top-level components: \texttt{reactants}, \texttt{products}, and \texttt{stages}. The \texttt{reactants} and \texttt{products} store detailed information for each molecular participant, while each \texttt{stage} represents a reaction step with associated conditions such as \texttt{reagent}, \texttt{catalyst}, and \texttt{solvent}. The schema covers all essential reaction information, including SMILES strings, reactant amounts, and product yields (Appendix~\ref{app:A}). This design is both semantically complete and practically useful, supporting downstream AI-for-Chemistry tasks such as product prediction~\cite{joung2025electron}, yield estimation, condition recommendation~\cite{wang2024identifying}, and retrosynthetic planning~\cite{xiong2025bridging,deng2025rsgpt,xu2025unified}. Accordingly, RxnExBench evaluates whether VLMs can accurately extract complete schema-compliant reaction records from real-world \textit{Screen Table} and \textit{Substrate Scope} figures.

\subsection{Difficulty-Aware Image Collection and Curation}
\label{subsec:difficulty}

As illustrated in Figure \ref{fig:fig1_REB_example}, \textit{Screen Tables} and \textit{Substrate Scopes} are two prevalent yet distinct formats for presenting reaction data in chemical publications (see Appendix~\ref{app:A} for detailed descriptions). In practice, they exhibit substantial heterogeneity in visual layout, information density, and required inference complexity.

Based on systematic analysis of real-world figures, we propose a three-tier
difficulty taxonomy along two orthogonal axes—\textit{visual scope} (local vs.\
global perception) and \textit{reasoning depth} (literal extraction vs.\
chemistry-aware inference). Each level is defined by concrete, observable operations a reader must perform:

\begin{itemize}
    \item \textbf{Level 1: Natural Reading, Local Perception, Immediate Completion.}
    Extracting structured chemical reaction information from Level 1 figures follows standard human reading order: first parse the reaction template and populate known information into the schema fields, then sequentially process each row (for \textit{Screen Table}) or substrate entry (for \textit{Substrate Scope}) to update fields such as specific reactants, catalysts, temperature, or time---thereby instantiating complete chemical reactions, as shown in Figure~\ref{fig:fig1_REB_multi_rxn}(a). This requires basic document understanding, OCR, molecular diagram recognition, and foundational chemical knowledge.

    \item \textbf{Level 2: Active Search, Global Perception, Integrated Reasoning.}  For Level~2 figures, linear reading is insufficient. Models must actively search and integrate information across distant regions---for example, when key conditions are specified only in footnotes, as illustrated by the purple arrow in Stage~2 of Figure~\ref{fig:fig1_REB_reasoning_stages}. Another challenge arises when a row or substrate corresponds to multiple reactions. We define figures with one reaction per row/substrate as ``simple-line-substrate'' (\textsf{SLS}) figures, and those containing multiple reactions as ``complex-line-substrate'' (\textsf{CLS}) figures (Figure~\ref{fig:fig1_REB_multi_rxn}(b)). These \textsf{CLS} figures belong to Level~2 or higher, requiring cross-region awareness and contextual integration.

    \item \textbf{Level 3: Expert Comprehension and Spatial Reasoning.}
    Building on Level 2, Level 3 requires domain-specific chemical reasoning. As shown in the Stage 3 of Figure~\ref{fig:fig1_REB_reasoning_stages}, in \textit{Substrate Scopes} (and complex \textit{Screen Tables}), the model must infer variable R-groups from substrate structures and propagate these substitutions back to reactants to reconstruct full molecular identities: a non-trivial spatial and symbolic inference task.
\end{itemize}

To construct a representative dataset, we collected $\sim$5,000 CC-4.0-licensed papers from 8 high-impact chemistry journals (list in Appendix~\ref{app:B}). After PDF-to-image conversion and chart region extraction using DocLayout-YOLO, we obtained $\sim$370K candidate figures. A custom classifier then identified 2,269 \textit{Substrate Scope} and 2,413 \textit{Screen Table} images as our initial pool.

From this large-scale collection, chemistry experts performed multi-round manual screening, preliminary annotation, and difficulty assessment to select the most representative and reasoning-intensive examples. The final benchmark contains 90 \textit{Substrate Scope} and 176 \textit{Screen Table} figures with coverage across all three difficulty levels (Table~\ref{tab:reaction-stats}).

\subsection{Data Annotation Pipeline}
\label{subsec:annotation}

Fully annotating a single \textit{Screen Table} or \textit{Substrate Scope} still requires an average of 230 and 339 minutes\footnote{Detailed statistics in Appendix~\ref{app:C.1}} respectively—reflecting both the deep chemical knowledge required and the visual parsing burden, and motivating our decomposition into focused, independently quality-controlled sub-tasks.
We therefore design a modular multi-stage annotation pipeline (Figure~\ref{fig:fig1_Data_Collection_and_Annotation}) that decomposes the workflow into three primary stages and several focused sub-tasks with independent quality control (details in Appendix~\ref{app:C.2}), improving both annotation efficiency and reliability:

\begin{enumerate}
    \item \textbf{Reaction Template Annotation}:
    Annotate all directly observable information in the generic reaction template (e.g., fixed reactants, base conditions). Fields requiring instance-specific context, such as R-group identities, yields, temperatures, or SMILES are left blank.

    \item \textbf{Template Instantiation}:
    For each substrate (\textit{Substrate Scope}) or table row (\textit{Screen Table}), instantiate the abstract template into a concrete reaction by filling in context-dependent fields (e.g., specific catalyst, solvent, yield). SMILES strings are still deferred.

    \item \textbf{Molecular Structure (SMILES) Annotation}:
   Relying on the early-stage model from~\cite{yang2026mineru} and the open-source tool \texttt{Ketcher}, chemists reconstruct complete molecular structures from diagrams and substrate context information, addressing the challenges posed by complex long-form SMILES representations and implicit R-group reasoning that existing diagram-to-SMILES models are unable to handle. The resulting \texttt{molfile} is then automatically converted into a standardized SMILES format using \texttt{RDKit}.
\end{enumerate}

We employed 40 chemistry-trained annotators with three rounds of cross-validation. In total, we invested 1,183 person-hours and produced 3,565 annotated reactions and 2,361 canonical SMILES strings. For more annotation details, please refer to Appendix~\ref{app:C}.

\begin{table*}[htbp]
    \centering
    \scriptsize
    \renewcommand{\arraystretch}{1.3}
  \caption{F1 scores of VLMs on RxnExBench (precision/recall in Appendix G). Results are shown for the full dataset and three difficulty levels under three evaluation criteria. \textbf{C} refers to \textbf{Criterion}. \textbf{Boldfaced} numbers denote the best performance, while \underline{underlined} numbers represent the runner-up.}
    \newcolumntype{C}{>{\centering\arraybackslash}X}
      \begin{tabularx}{\textwidth}{p{8.5em} | *{3}{C} | *{3}{C} | *{3}{C} | *{3}{C}} % 把第一列加宽到8.5em
      \toprule
      \multirow{2}{*}{\textbf{Models}}
      & \multicolumn{3}{c|}{\textbf{All Levels (266 Figures)}}
      & \multicolumn{3}{c|}{\textbf{Level 1 (60 Figures)}}
      & \multicolumn{3}{c|}{\textbf{Level 2 (90 Figures)}}
      & \multicolumn{3}{c}{\textbf{Level 3 (116 Figures)}} \\
      & C1 & C2 & C3
      & C1 & C2 & C3
      & C1 & C2 & C3
      & C1 & C2 & C3 \\
      \midrule
      \rowcolors{1}{gray!10}{white}
      Claude Opus 4.6 & 2.52\% & 1.08\% & 0.94\% & 5.60\% & 4.24\% & 4.24\% & 2.40\% & 1.20\% & 0.74\% & 1.62\% & 0.00\% & 0.00\% \\
      Gemini 3.1 Pro & 19.47\% & \textbf{9.57\%} & \textbf{8.86\%} & \textbf{40.42\%} & \textbf{20.56\%} & \textbf{19.86\%} & 15.46\% & \underline{10.22\%} & \textbf{8.25\%} & \underline{15.24\%} & \textbf{5.77\%} & \textbf{5.77\%} \\
      GPT-5.5 & \textbf{23.16\%} & \underline{9.16\%} & \underline{6.81\%} & 33.50\% & \underline{12.79\%} & \underline{12.63\%} & \textbf{28.13\%} & \textbf{13.93\%} & \underline{6.41\%} & \textbf{16.91\%} & \underline{5.18\%} & \underline{5.18\%} \\
      GPT-4o & 0.09\% & 0.00\% & 0.00\% & 0.00\% & 0.00\% & 0.00\% & 0.00\% & 0.00\% & 0.00\% & 0.17\% & 0.00\% & 0.00\% \\
      Doubao Seed 2.0 Pro & \underline{19.81\%} & 6.59\% & 5.35\% & \underline{33.56\%} & 11.36\% & 10.68\% & \underline{25.70\%} & 7.14\% & 4.55\% & 11.80\% & 4.71\% & 4.10\% \\
      GLM-5V Turbo & 5.08\% & 1.92\% & 1.80\% & 15.48\% & 6.80\% & 6.29\% & 2.77\% & 1.29\% & 1.20\% & 2.90\% & 0.60\% & 0.60\% \\
      Intern-S1 & 2.88\% & 0.06\% & 0.06\% & 4.59\% & 0.00\% & 0.00\% & 6.90\% & 0.19\% & 0.19\% & 0.00\% & 0.00\% & 0.00\% \\
      Llama 4 Maverick & 2.27\% & 0.86\% & 0.86\% & 4.58\% & 2.38\% & 2.38\% & 2.83\% & 0.10\% & 0.10\% & 0.99\% & 0.80\% & 0.80\% \\
      Qwen3-VL-235B & 1.68\% & 0.26\% & 0.06\% & 5.62\% & 1.53\% & 0.34\% & 2.23\% & 0.00\% & 0.00\% & 0.00\% & 0.00\% & 0.00\% \\
      \bottomrule
    \end{tabularx}
     \label{tab:all_result}
  \end{table*}

\subsection{Evaluation Task Design}
\label{subsec:evaluation}

We formulate the task as end-to-end extraction: given a \textit{Screen Table} or \textit{Substrate Scope} image, the VLM must output \textit{all} reactions in the image as a single JSON object compliant with the schema (prompt template in Appendix~\ref{app:E}).

We evaluate using \textit{reaction-level} precision, recall and F1. For each image, predicted and ground-truth reaction lists are matched, and three criteria determine whether a prediction is a true positive (TP):

\begin{itemize}
    \item \textbf{Criterion 1} (for retrosynthesis, etc.): match only on \texttt{reactants}, \texttt{products}, and \texttt{reagents}.
    \item \textbf{Criterion 2} (for yield prediction): adds \texttt{catalyst} and \texttt{solvent} to Criterion 1.
    \item \textbf{Criterion 3} (for condition recommendation, etc.): the strictest standard, requiring exact match on \texttt{reactants}, \texttt{products}, \texttt{reagents}, \texttt{catalyst}, \texttt{ligand}, \texttt{solvent}, \texttt{pressure}, \texttt{reaction time}, \texttt{reaction temperature}, and \texttt{pH}.
\end{itemize}

A prediction is counted as TP \textit{only if all fields specified by the criterion match exactly}; otherwise, it is FP or FN. Full evaluation protocol is detailed in Appendix~\ref{app:F}.

\section{Experiments}
\label{sec:experiments}

\subsection{Model Selection and Experimental Setup}
\label{subsec:model_selection}

We conducted an in-depth evaluation of state-of-the-art Vision-Language Models (VLMs), including both open-source models and commercial APIs: Claude~Opus~4.6~\cite{anthropic2025website}, Gemini~3.1~Pro~\cite{team2023gemini}, GPT-5.5~\cite{openai2025gptoss}, GPT-4o~\cite{hurst2024gpt}, Doubao~Seed~2.0~Pro~\cite{guo2025seed1}, GLM-5V~Turbo~\cite{vteam2025glm45vglm41vthinkingversatilemultimodal}, Intern-S1~\cite{bai2025intern}, Llama~4~Maverick~\cite{llama2025website}, and Qwen3-VL-235B~\cite{qwen2025automated}.

\subsection{Overall Performance and Preliminary Findings}
\label{subsec:overall_performance}

The F1 scores of all models across different difficulty levels and evaluation criteria on RxnExBench are presented in Table~\ref{tab:all_result}, with precision and recall metrics detailed in Appendix~\ref{app:G}. Our analysis reveals three key findings:

\textbf{First, complete and accurate extraction of structured chemical reaction information from \textit{Screen Tables} and \textit{Substrate Scopes} remains highly challenging for state-of-the-art VLMs}. Under Criterion~1—the most practically relevant setting for retrosynthesis-style downstream tasks—GPT-5.5 leads at 23.16\% F1, with Doubao~Seed~2.0~Pro (19.81\%) and Gemini~3.1~Pro (19.47\%) forming a clear second tier. The remaining six models cluster below 6\%. Under the stress-test Criterion~3,
which requires exact matching across ten fields, even the strongest model reaches only 8.86\% F1. Together, these results highlight substantial room for improvement in current VLMs for structured chemical information extraction. Human expert performance on the same task is reported in Appendix~\ref{app:C.5}.

\begin{figure}[!t]
    \centering
    \includegraphics[width=0.8\linewidth]{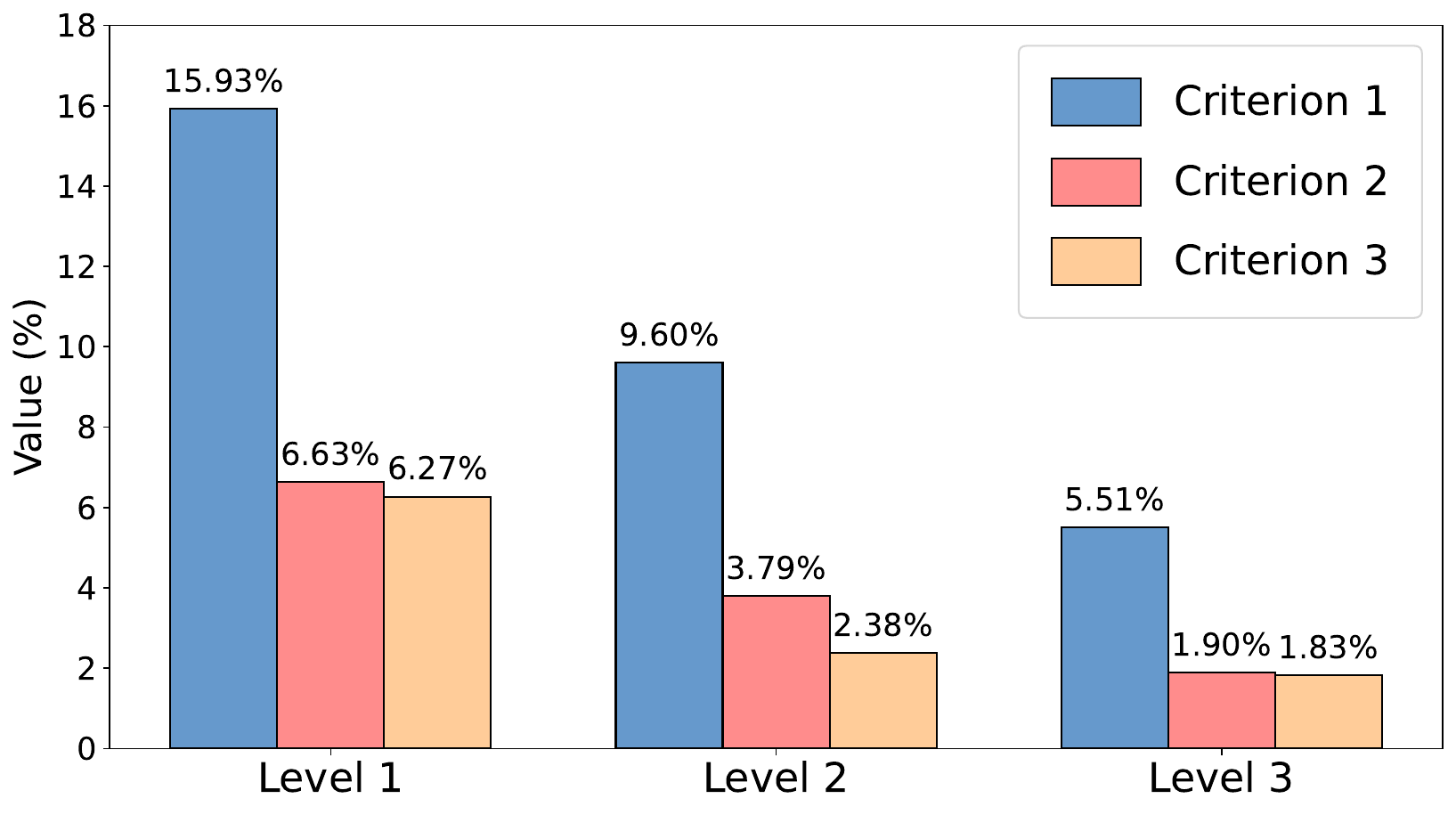}
    \caption{Average F1 scores of all VLMs across RxnExBench difficulty levels and evaluation criteria.}
    \label{fig:REB_Criterions_comparison}
    \vspace{-1.0\baselineskip}
\end{figure}

\textbf{Second, extraction difficulty increases significantly when downstream tasks require more fields from the \textit{Schema}}. As shown in Figure~\ref{fig:REB_Criterions_comparison}, when progressing from Criterion~1 to Criterion~3, all VLMs exhibit a consistent and significant performance decline, demonstrating that field complexity is a critical factor affecting extraction accuracy.

\textbf{Third, VLM performance varies substantially across difficulty levels}. Figure~\ref{fig:REB_Criterions_comparison} illustrates the average performance distribution of all 9 models across the three difficulty levels, revealing a clear stepwise performance degradation from Level 1 to Level 3. For instance, Gemini~3.1~Pro achieves only 15.24\% F1 on Level~3 data under Criterion~1, significantly lower than that on Level~1 data under the same criterion (40.42\%).

\subsection{Reaction Count as a Proxy for Basic Understanding Ability}
\label{subsec:reaction_count_proxy}

Beyond aggregate F1, we introduce \textit{reaction-count consistency} as a complementary diagnostic that targets a more fundamental ability: whether a VLM correctly perceives \emph{how many} reactions a figure contains, independently of whether each individual reaction is extracted correctly. Because every downstream field is bound to a specific reaction, miscounting reactions propagates errors into the entire output; accurate reaction counting is therefore a prerequisite for any structured extraction and a clean probe of basic chart parsing, decoupled from chemistry-aware reasoning.

As defined in Section~\ref{subsec:difficulty}, RxnExBench figures are
labeled as \textsf{SLS}  or \textsf{CLS} . We analyze the two groups separately. For \textsf{SLS} figures, VLMs can generally match the ground truth reaction count, indicating basic chart parsing ability. However, for \textsf{CLS} figures, performance drops sharply: even the best model (Doubao 2.0) achieves only 60\% consistency with ground truth, with most VLMs significantly undercounting reactions, demonstrating that multi-reaction rows require broader context awareness beyond natural reading order.

\begin{figure}[!t]
    \centering
    \includegraphics[width=0.8\linewidth]{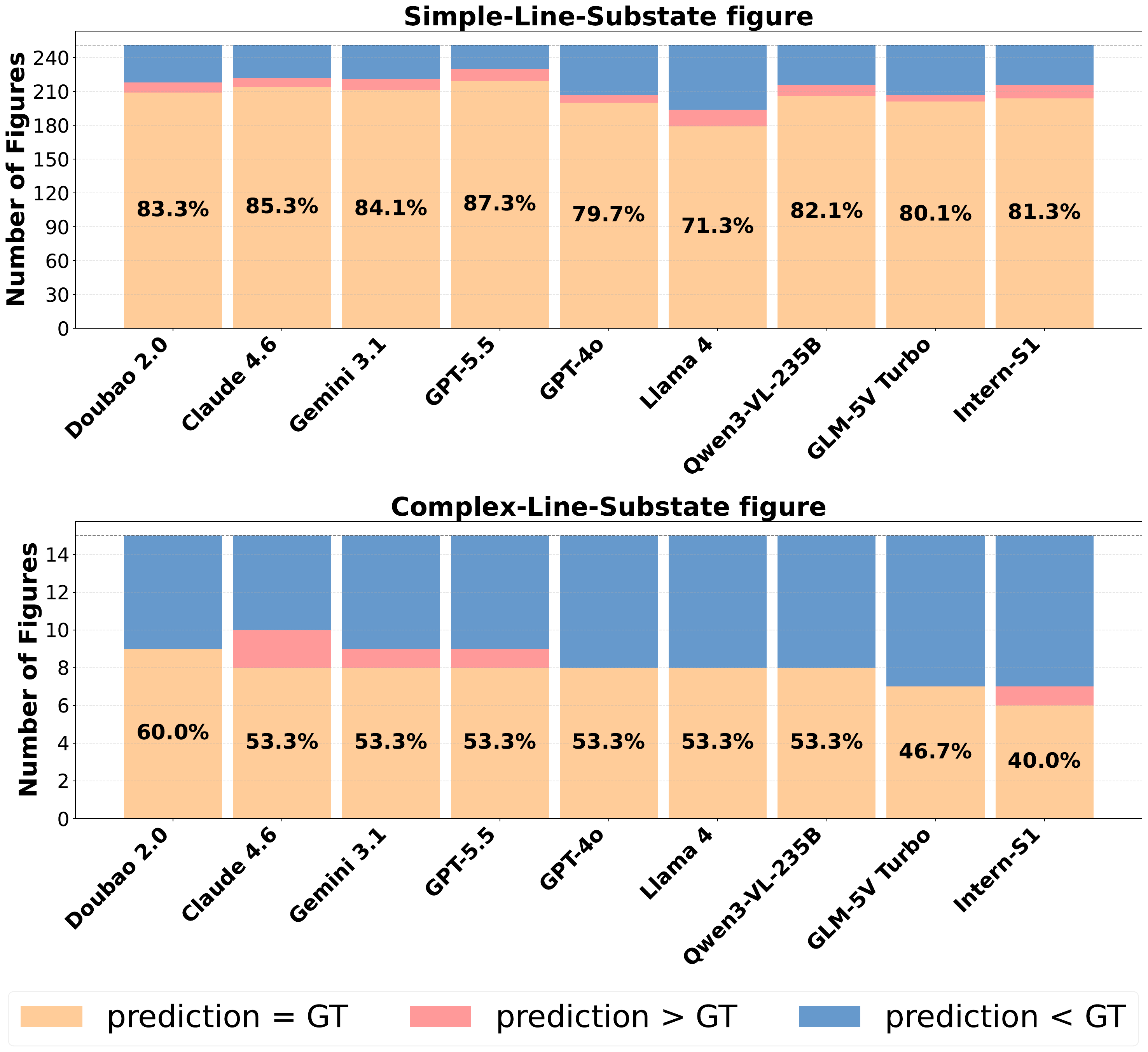}
    \caption{Comparison of reaction counts in figure between VLM predictions and ground truth, grouped by simple-line-substrate (SLS) and complex-line-substrate (CLS) figures. While VLMs achieve accurate reaction counts on SLS figures, they exhibit substantial omission errors on CLS figures.}
    \label{fig:REB_RXN_NUMS}
\end{figure}

\subsection{Error Analysis}
\label{subsec:error_analysis}

To systematically analyze VLM failure modes, we categorize errors into three non-exclusive types: \textbf{Field Missing} (required fields omitted), \textbf{SMILES Error} (incorrect SMILES for molecules drawn as diagrams), and \textbf{Content Error} (incorrect content for non-SMILES fields).

\begin{figure}[!t]
    \centering
    \includegraphics[width=1\linewidth]{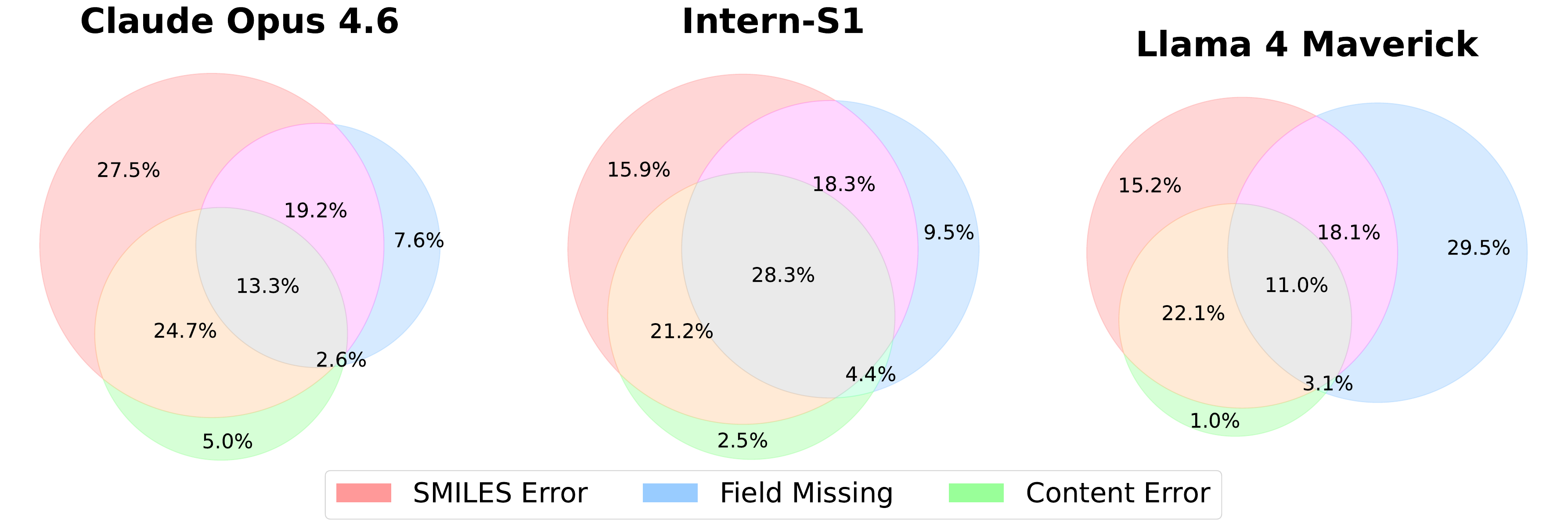}
    \caption{Error type distributions for three representative models.}
    \label{fig:fig1_REB_venn_diagrams}
    \vspace{-1.0\baselineskip}
\end{figure}

We performed systematic error analysis for all VLMs on RxnExBench (detailed results in Appendix~\ref{app:G}). SMILES-related errors dominate the failure modes: 75.51\% of extraction errors involve incorrect molecular structures, highlighting molecular recognition as a major bottleneck in automated chemical reaction extraction.

However, the distribution of error types varies substantially across models (Figure~\ref{fig:fig1_REB_venn_diagrams}). For Claude~Opus~4.6, 84.7\% of erroneous reactions involve SMILES errors, with 27.5\% arising exclusively from molecular recognition, indicating that molecular structure recognition is its primary failure mode. In contrast, Intern-S1 exhibits strong overlap among all three error categories, suggesting simultaneous weaknesses in visual perception, chemical reasoning, and molecular recognition. Llama~4~Maverick shows a different pattern: Field Missing errors form the largest independent component (29.5\%), largely because the model often fails to identify complete reactions (Figure~\ref{fig:REB_RXN_NUMS}).

Beyond these model-specific patterns, we observe two consistent trends. First, models that perform well under Criterion~1 but degrade sharply under Criterion~3 (e.g., Gemini~3.1~Pro and GPT-5.5) frequently identify reactants and products correctly while missing or merging condition fields, indicating that condition extraction becomes a major challenge under stricter evaluation. Second, all models systematically undercount reactions on \textsf{CLS} figures (Section~\ref{subsec:reaction_count_proxy}), revealing weaknesses in layout-aware reasoning when a single row corresponds to multiple reactions. Together, these findings identify molecular recognition, condition extraction, and layout reasoning as three coupled bottlenecks for current VLMs.
\begin{table}[t]
    \centering
    \scriptsize
    \setlength{\tabcolsep}{3pt}
    \renewcommand{\arraystretch}{1.1}
    \caption{VLM performance (F1 scores) under \textsf{Oracle-MolRec inference} setting on the \textbf{RxnExBench-MolIdt} subset. (\textcolor{red}{Red numbers in parentheses}) show improvement over the \textsf{standard inference} setting.}
    \resizebox{0.8\columnwidth}{!}{%
    \begin{tabular}{l | ccc}
    \toprule
    \textbf{Models} & Criterion 1 & Criterion 2 & Criterion 3 \\

    \midrule

    Claude Opus 4.6
    & 30.04\%(\textcolor{red}{↑27.52\%})
    & \underline{17.20\%}(\textcolor{red}{↑16.12\%})
    & \underline{14.78\%}(\textcolor{red}{↑13.84\%}) \\

    Gemini 3.1 Pro
    & 36.70\%(\textcolor{red}{↑17.23\%})
    & \textbf{18.25\%}(\textcolor{red}{↑8.68\%})
    & \textbf{15.88\%}(\textcolor{red}{↑7.02\%}) \\

    GPT-5.5
    & \textbf{42.67\%}(\textcolor{red}{↑19.51\%})
    & 16.83\%(\textcolor{red}{↑7.67\%})
    & \underline{15.29\%}(\textcolor{red}{↑8.48\%}) \\

    GPT-4o
    & 25.25\%(\textcolor{red}{↑25.16\%})
    & 5.32\%(\textcolor{red}{↑5.32\%})
    & 4.57\%(\textcolor{red}{↑4.57\%}) \\

    Doubao Seed 2.0 Pro
    & \underline{36.73\%}(\textcolor{red}{↑16.92\%})
    & 11.94\%(\textcolor{red}{↑5.35\%})
    & 10.41\%(\textcolor{red}{↑5.06\%}) \\

    GLM-5V Turbo
    & 31.95\%(\textcolor{red}{↑26.87\%})
    & 11.99\%(\textcolor{red}{↑10.07\%})
    & 11.11\%(\textcolor{red}{↑9.31\%}) \\

    Intern-S1
    & 20.77\%(\textcolor{red}{↑17.89\%})
    & 7.41\%(\textcolor{red}{↑7.35\%})
    & 5.87\%(\textcolor{red}{↑5.81\%}) \\

    Llama 4 Maverick
    & 20.34\%(\textcolor{red}{↑18.07\%})
    & 9.08\%(\textcolor{red}{↑8.22\%})
    & 8.25\%(\textcolor{red}{↑7.39\%}) \\

    Qwen3-VL-235B
    & 26.90\%(\textcolor{red}{↑25.22\%})
    & 7.24\%(\textcolor{red}{↑6.98\%})
    & 5.94\%(\textcolor{red}{↑5.88\%}) \\

    \bottomrule
    \end{tabular}%
    }

    \label{tab:smiles_wgt}
    \vspace{-1.0\baselineskip}
\end{table}

\subsection{Disentangling Molecular Recognition from Higher-Level Reasoning}
\label{subsec:ablation_mol_rec}

Since molecular recognition (OCSR) is a well-established task with many specialized models, integrating external OCSR systems into the extraction pipeline could alleviate a major bottleneck. Inspired by~\cite{song2026rxncaption,song2026molecular}, to estimate the upper bound of this approach, we introduce the \textsf{Oracle-MolRec} protocol, which provides ground-truth SMILES for every directly drawn molecule together with references linking each SMILES to its corresponding structure in the image, thereby simulating perfect molecular recognition. We instantiate three variants: \textsf{Oracle-MolRec-IDT}, which pairs molecular identifiers with SMILES (default setting); \textsf{Oracle-MolRec-BBOX-IDX}, which annotates molecules with indexed bounding boxes and provides index--SMILES mappings; and \textsf{Oracle-MolRec-BBOX-CORD}, which supplies BBOX-coordinate--SMILES pairs. Prompt templates are provided in Appendix~\ref{app:E}.

Table~\ref{tab:smiles_wgt} compares \textsf{Oracle-MolRec-IDT} against standard inference. Performance improves substantially across all criteria, confirming the importance of accurate molecular structures for reaction extraction. Under Criterion~1, the best F1 score increases from 23.16\% to 42.67\%; however, performance under the stricter Criteria~2 and~3 remains limited, reaching only 18.25\% and 15.88\%, respectively. Notably, under Criterion~3, 4 of the 9 evaluated VLMs still achieve below 10\% F1 even with perfect molecular recognition.

These results show that parsing \textit{Screen Tables} and \textit{Substrate Scopes} requires far more than localized molecular recognition. Beyond accurate OCSR, successful extraction depends on holistic multimodal perception, long-range integration of templates, instances, and footnotes, and chemistry-aware reasoning. In this sense, \textsf{Oracle-MolRec} serves not only as an oracle baseline, but also as a probe for disentangling molecular recognition from higher-level reasoning. Similar gain and plateau patterns are consistently observed under both \textsf{Oracle-MolRec-BBOX-IDX} and \textsf{Oracle-MolRec-BBOX-CORD}; detailed comparisons are provided in Appendix~\ref{app:G}.

\section{Conclusion}

We introduce RxnExBench,  the first benchmark that jointly covers \textit{Screen Tables} and \textit{Substrate Scopes} with full reaction-level schema, R-group resolution, CLS figures, and a reasoning-driven difficulty taxonomy. The benchmark contains 3,565 expert-annotated reactions and 2,361 molecular structures across three reasoning-driven difficulty levels under a 26-field reaction schema.

Experiments on 9 state-of-the-art VLMs show that current models still struggle with this task, particularly in multimodal integration and chemistry-aware reasoning. Oracle-MolRec results further reveal that perfect molecular recognition alone is insufficient for accurate reaction extraction. We hope RxnExBench will support future research in automated reaction extraction and AI-for-Chemistry.

\section*{Limitations}

Although RxnExBench provides a challenging and high-quality benchmark for evaluating Vision-Language Models on structured chemical reaction extraction, the current dataset is still primarily focused on English-language organic synthesis literature and adopts relatively strict reaction-level matching criteria. Therefore, the reported results should be viewed as a snapshot of the current capabilities of general-purpose VLMs rather than the upper bound of task performance. In the future, RxnExBench can be further expanded to cover broader chemistry sub-disciplines and literature sources, while also enabling the exploration of supervised fine-tuning, tool-augmented approaches, and related strategies to further advance this field.

\section*{Acknowledgments}

This work is supported by Shanghai Artificial Intelligence Laboratory.

\clearpage
\bibliographystyle{plainnat}
\bibliography{paper}

\clearpage
\beginappendix

\section{Chemical Reaction Related Concepts and Reaction Schema}
\label{app:A}
\subsection{Introduction of Screen Tables and Substrate Scopes}

\begin{figure}[h]
    \centering
    \includegraphics[width=1\linewidth]{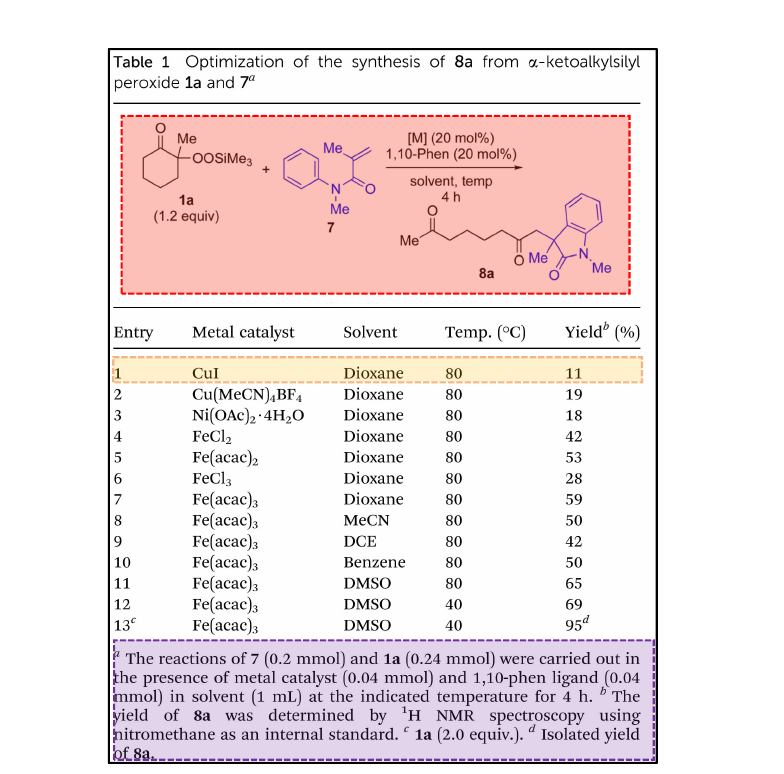}
    \caption{Example of Screen Table.}
    \label{fig:App_a1_table}
\end{figure}

As shown in Figure \ref{fig:App_a1_table}, \textit{Screen Table} is a common form of table in chemical papers, used to systematically compare the performance of a set of highly similar chemical reactions under strictly controlled variables (such as different catalysts, solvents, temperatures, reaction times, etc.), thereby identifying the optimal reaction conditions.

A typical \textit{Screen Table} usually includes a ``reaction template'' presented as an image (as indicated by the red box in Figure \ref{fig:App_a1_table}), which summarizes the core information shared by the entire set of reactions. This includes reactants and products (typically represented as molecular structure diagrams) and some reaction conditions (often annotated as text). In the template, certain conditions are explicitly specified (such as the reagent ``1,10-Phen'' labeled in Figure \ref{fig:App_a1_table}), while others are represented by abstract symbols (e.g., ``solvent/temp'' above the reaction arrow).

Below the template, a structured table is usually attached, where each row (i.e., an \textit{entry}) corresponds to the reaction results under one or more specific experimental conditions. However, the information in a single row is often incomplete and must be combined with the reaction template above to reconstruct a complete reaction description. Common scenarios include: instantiating abstract conditions from the template (e.g., replacing ``solvent/temp'' with specific compound names), or making localized modifications to a condition in the template (e.g., explicitly setting the reaction temperature to ``80\,$^\circ$C'' in row 1).

Additionally, some  \textit{Screen Tables} include footnotes at the bottom of the table (as shown in the purple box in Figure \ref{fig:App_a1_table}) to supplement experimental details or special notes that cannot be fully expressed in the table.

\begin{figure}[h]
    \centering
    \includegraphics[width=1\linewidth]{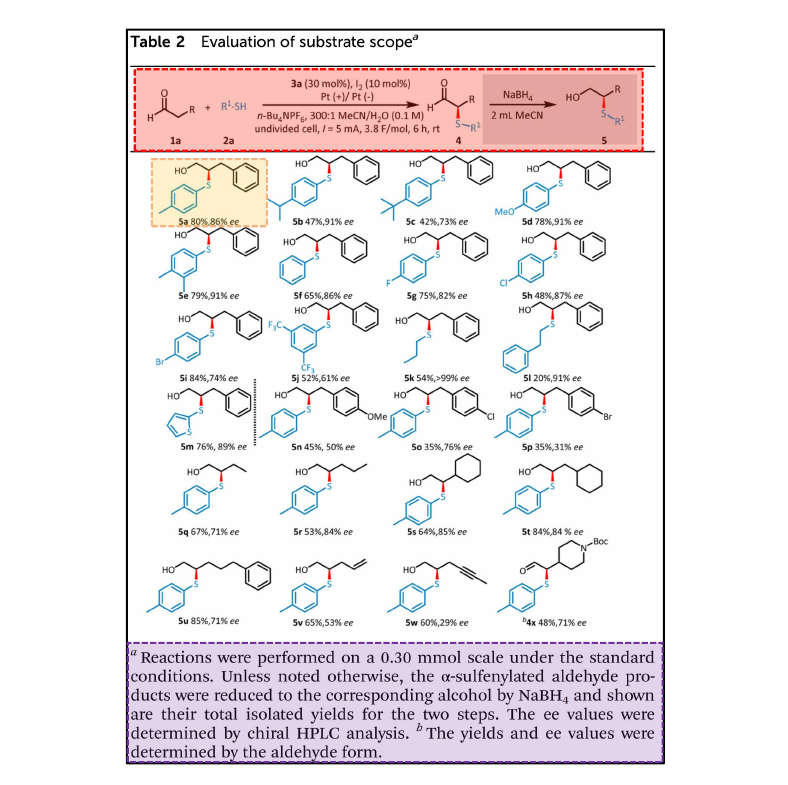}
    \caption{Example of Substrate Scope.}
    \label{fig:App_a1_scope}
\end{figure}

Similar to  \textit{Screen Table}, \textit{Substrate Scope} (as shown in Figure \ref{fig:App_a1_scope}) also includes a reaction template in the form of an image. The key difference is that in \textit{Substrate Scope}, the reaction conditions are typically fixed and explicitly defined, while the molecular structures of the reactants and products contain one or more variable substituents (such as R and R\textsuperscript{1} in Figure \ref{fig:App_a1_scope}; other common forms include R$'$ or specific symbols).

Below the reaction template, molecular structure diagrams of specific ``substrates'' and their corresponding reaction results (such as yield, selectivity, etc.) are neatly arranged. These substrate structures (e.g., \textbf{5a–5w} in Figure \ref{fig:App_a1_scope}) essentially represent ``instantiations'' of the product in the template (such as molecule \textbf{5}); they clarify the exact form of the product generated when the substrate participates in the reaction by specifying the specific chemical structure of the R-group.

Similar to \textit{Screen Tables}, \textit{Substrate Scopes} often include footnotes below the substrate array to supplement critical experimental details that are difficult to present directly in the diagram (as shown in the purple box in Figure \ref{fig:App_a1_scope}) .

% Table generated by Excel2LaTeX from sheet 'Sheet2'
\begin{table}[t]
  \centering
  \scriptsize
  \caption{Structured chemical reaction schema of RxnExbench.}
    \begin{tabular}{c|p{5em}|p{11em}}
    \toprule
    \multicolumn{2}{p{15em}|}{\textbf{Reaction roles}} & \textbf{Fine grained fields} \\
    \midrule

    % reactants
    \multicolumn{2}{p{15em}|}{reactants (list)}
    & SMILES (str) \newline{}text (str)\newline{}amount (str)\newline{}conversion rate (str) \\
    \midrule

    % products
    \multicolumn{2}{p{15em}|}{\multirow{2}{*}{products (list)}}
    & SMILES (str) \newline{}text (str) \newline{}yield ratio (str)
    \newline{}isolated yield (str)\newline{}enantiomeric excess (str)\newline{}regioisomeric ratio (str) \newline{}enantiomeric ratio (str) \newline{}alkene configuration (str) \newline{}diastereomeric ratio (str) \\
    \multicolumn{2}{p{15em}|}{} & \\
    \midrule

    % stage
    \multicolumn{1}{c|}{\multirow{14}{*}{stage (list)}}

    & \multicolumn{1}{p{12em}|}{\multirow{2}{*}{reagents (list)}}
    & SMILES (str) \newline{}text (str) \newline{}amount (str) \\
    & \multicolumn{1}{p{12em}|}{} & \\
\cmidrule{2-3}

    & \multicolumn{1}{p{12em}|}{catalysts (list)}
    & SMILES (str) \newline{}text (str) \newline{}amount (str) \\
\cmidrule{2-3}

    & \multicolumn{1}{p{12em}|}{solvents (list)}
    & text (str) \newline{}ratio (str) \newline{}amount (str) \\
\cmidrule{2-3}

    & \multicolumn{1}{p{12em}|}{ligand (list)}
    & SMILES (str) \newline{}text (str) \newline{}amount (str) \\
\cmidrule{2-3}

    & \multicolumn{1}{p{12em}|}{pressure (str)}
    & - \\
\cmidrule{2-3}

    & \multicolumn{1}{p{12em}|}{reaction time (str)}
    & - \\
\cmidrule{2-3}

    & \multicolumn{1}{p{12em}|}{reaction temperature (str)}
    & - \\
\cmidrule{2-3}

    & \multicolumn{1}{p{12em}|}{light condition (str)}
    & - \\
\cmidrule{2-3}

    & \multicolumn{1}{p{12em}|}{atmosphere (str)}
    & - \\
\cmidrule{2-3}

    & \multicolumn{1}{p{12em}|}{pH (str)}
    & - \\
\cmidrule{2-3}

    & \multicolumn{1}{p{12em}|}{stir speed (str)}
    & - \\
\cmidrule{2-3}

    & \multicolumn{1}{p{12em}|}{vacuum condition (str)}
    & - \\
\cmidrule{2-3}

    & \multicolumn{1}{p{12em}|}{scale (str)}
    & - \\
    \bottomrule
    \end{tabular}%
  \label{tab:Schma}%
\end{table}%

\subsection{Structured Chemical Reaction Schema}
We define a structured schema for representing chemical reactions. Table \ref{tab:Schma} presents the detailed specifications of the Structured Reaction Schema. Since some reaction roles or fields may contain multiple instances, we have introduced the corresponding ``list'' data type in addition to the ``string'' type. The data types are indicated in parentheses. Additionally, the fields within each reaction role are organized in a ``dictionary'' format. Specific annotation examples are shown in Appendix~\ref{app:D}.

\begin{table}[t]
\centering
\caption{Journal List}
\begin{tabular}{l}
\toprule
\textbf{Journal Name} \\
\midrule
Angewandte Chemie International Edition \\
Chemical Science \\
Journal of the American Chemical Society \\
ACS Catalysis \\
Green Chemistry \\
ACS Central Science \\
Nature Chemistry \\
Accounts of Chemical Research \\
\bottomrule
\end{tabular}
\label{tab:journal_list}
\end{table}

\section{Data Collection}
\label{app:B}
Our data come from 8 high-impact chemistry journals. The list of journals is shown in Table \ref{tab:journal_list}.
\begin{table}[H]
\small
\centering
\caption{Average Per-image Annotation Statistics}
\label{tab:reaction_stats}
\begin{tabular}{lcc}
\toprule
\textbf{Figure Type} & \textbf{Screen Table} & \textbf{Substrate Scope} \\
\midrule
Reactions & 11.53 & 17.07 \\
Annotated Fields & 141.81 & 248.17 \\
Molecular structures & 36.17 & 43.67 \\
Atoms & 441.65 & 584.00 \\
Chemical bonds & 456.76 & 605.00 \\
\bottomrule
\end{tabular}
\label{tab:appendix_c_ann_stat}
\end{table}
\section{Data annotation}
\label{app:C}
\subsection{Statistics Related to Data Annotation}
\label{app:C.1}
The RxnExBench dataset comprises 266 images and 3,565 chemical reactions in total. During the annotation of reaction templates, 53,659 fields and 10,296 molecular structures were annotated. The 10,296 figure counts molecule \emph{occurrences}---a single molecule may be
referenced by multiple reactions; after RDKit canonicalization and deduplication, these correspond to the 2,361 unique canonical SMILES reported in the main text. Collectively, these structures contain 130,290 atoms and 134,840 chemical bonds. The average statistics per image are summarized in Table \ref{tab:appendix_c_ann_stat}.
\subsection{Detailed Data Annotation Process}
\label{app:C.2}
Our annotation pipeline consists of three main stages, as outlined in the main text: \textbf{Reaction Template Annotation}, \textbf{Template Instantiation}, and \textbf{Molecular Structure (SMILES) Annotation}. To provide necessary context and enable collaborative work, two preparatory steps are integrated into this workflow: \textbf{Molecular Structure Localization} precedes the Reaction Template Annotation, and the \textbf{Template Instantiation} stage is itself divided into two sub-steps: \textbf{Pre-annotation of Instantiation References} and \textbf{Instantiation Execution}.

\textbf{Molecular Structure Localization}. In this initial step, annotators are required to draw bounding boxes around all molecular structures depicted as diagrams (non-string form) within the image. This localization serves as a critical reference for subsequent stages: these diagram-based structures must be specifically annotated later using the \texttt{Ketcher} tool to generate their corresponding SMILES strings. The bounding boxes allow us to visually guide annotators to the correct molecules in later steps and to organize the molecular annotation content using the unique indices of these boxes.

\textbf{Reaction Template Annotation}. As described in the main text, annotators delineate the area of a generic reaction template using a polygon on the annotation platform. The initial annotation is performed based solely on the information contained within this polygon. Consistent with the main text, fields requiring instance-specific context, such as R-group identities, yields, temperatures, or SMILES, are left blank at this stage.

\textbf{Pre-annotation of Instantiation References}. A single image often contains one reaction template that must be instantiated multiple times. For instance, each row in a \textit{Screen Table} or each substrate molecule in a \textit{Substrate Scope} figure typically corresponds to one concrete reaction instance. Since our annotation process is collaborative, with multiple annotators working from a shared pool of images, different instantiation tasks for the same image might be handled by different individuals. To provide context, this sub-step requires annotators to first register all information that easily identifies each reaction instance across the image. This includes details like row numbers in a \textit{Screen Table} or the bounding box indices of substrate molecules in a \textit{Substrate Scope}. This registry of references is crucial for the subsequent execution of instantiation tasks.

\textbf{Instantiation Execution}. After the reference information for all instances within an image is established in the previous sub-step, a distinct instantiation task is created for each entry in this registry. Annotators, provided with a specific instantiation context (e.g., a row number, a substrate box index), proceed to convert the abstract template from the \textbf{Reaction Template Annotation} into a concrete reaction. They fill in the context-dependent fields (e.g., specific catalyst, solvent, yield). For molecular SMILES, only the ID of the corresponding molecular bounding box is recorded at this point; the actual SMILES string is added later.

\textbf{Molecular Structure (SMILES) Annotation}. In this step, annotators use the \texttt{Ketcher} tool to annotate specified molecular structures one at a time. The molecule to be annotated is highlighted using its pre-defined bounding box. For molecules within reaction templates that contain undefined R-groups—whose identities are determined only during template instantiation—the relevant instantiation reference information (from the pre-annotation sub-step) is also provided. This ensures the annotator knows exactly which reaction instance (e.g., which row or substrate) the R-group corresponds to, allowing for the correct structure to be drawn.

Finally, the \texttt{mol\_file} information for each molecule, obtained from the \textbf{Molecular Structure (SMILES) Annotation} step, is programmatically converted into a standardized SMILES string using \texttt{RDKit}~\cite{rdkit}. Based on the associated molecular box ID and the instantiation reference information, these SMILES strings are automatically inserted into their corresponding slots within the instantiated reaction templates from the \textbf{Instantiation Execution} step. This process yields the final, complete ground truth (GT) data.

\subsection{Data Annotation Platform}
\label{app:C.3}
\begin{figure}[!t]
    \centering
    \includegraphics[width=0.94\linewidth]{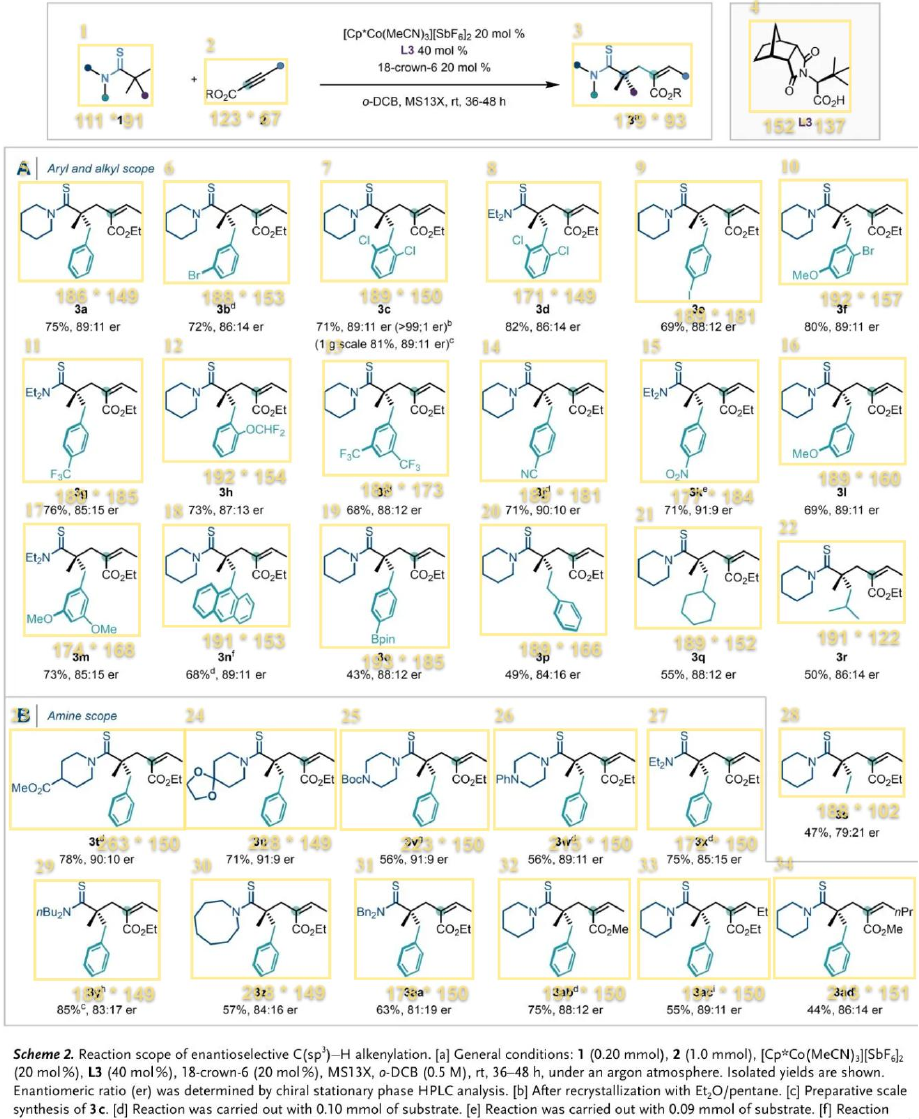}
    \caption{Screenshot for Molecular Structure Localization}
    \label{fig:appendix_c_mol_det}
\end{figure}

\begin{figure}[!t]
    \centering
    \includegraphics[width=0.94\linewidth]{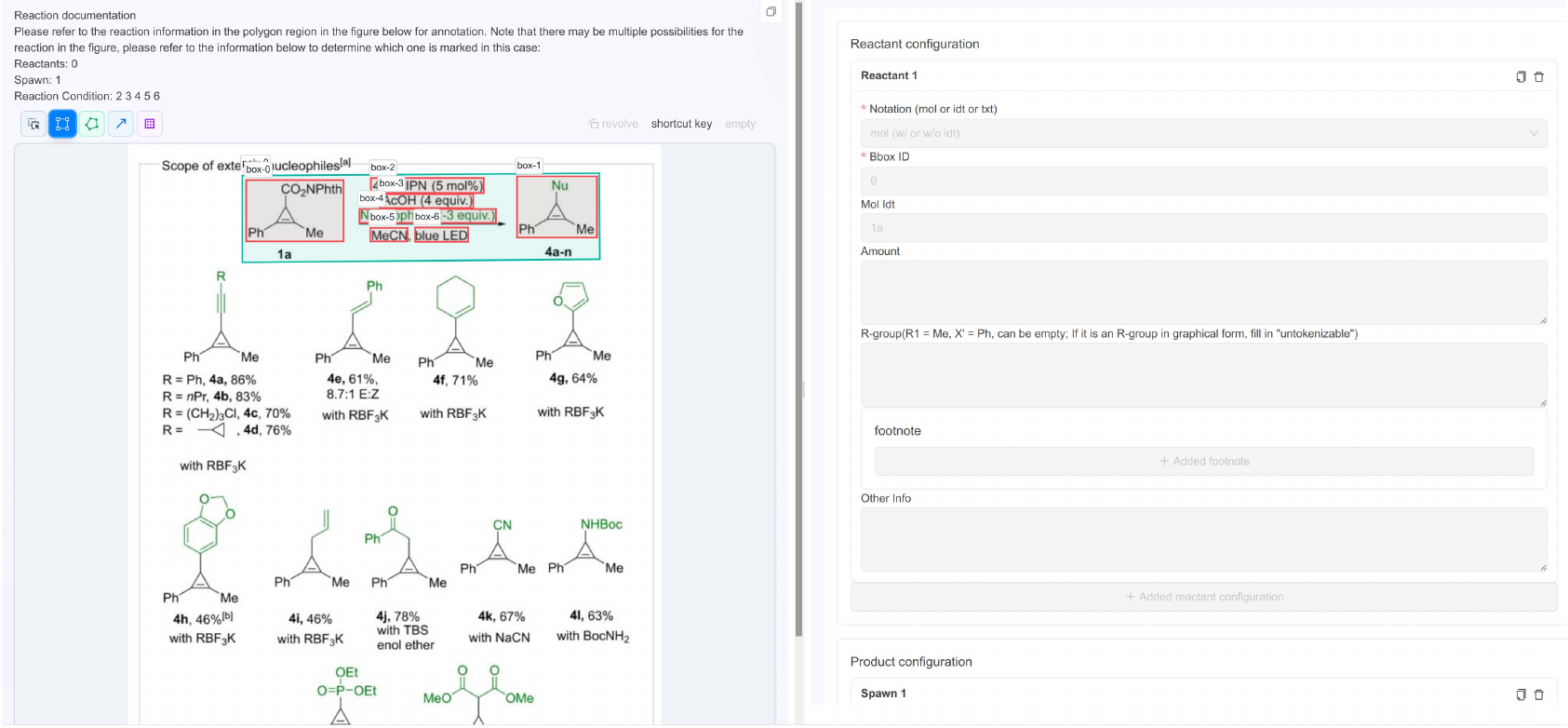}
    \caption{Screenshot for Reaction Template Annotation}
    \label{fig:appendix_c_rxn_abs}
\end{figure}

\begin{figure}[!t]
    \centering
    \includegraphics[width=0.94\linewidth]{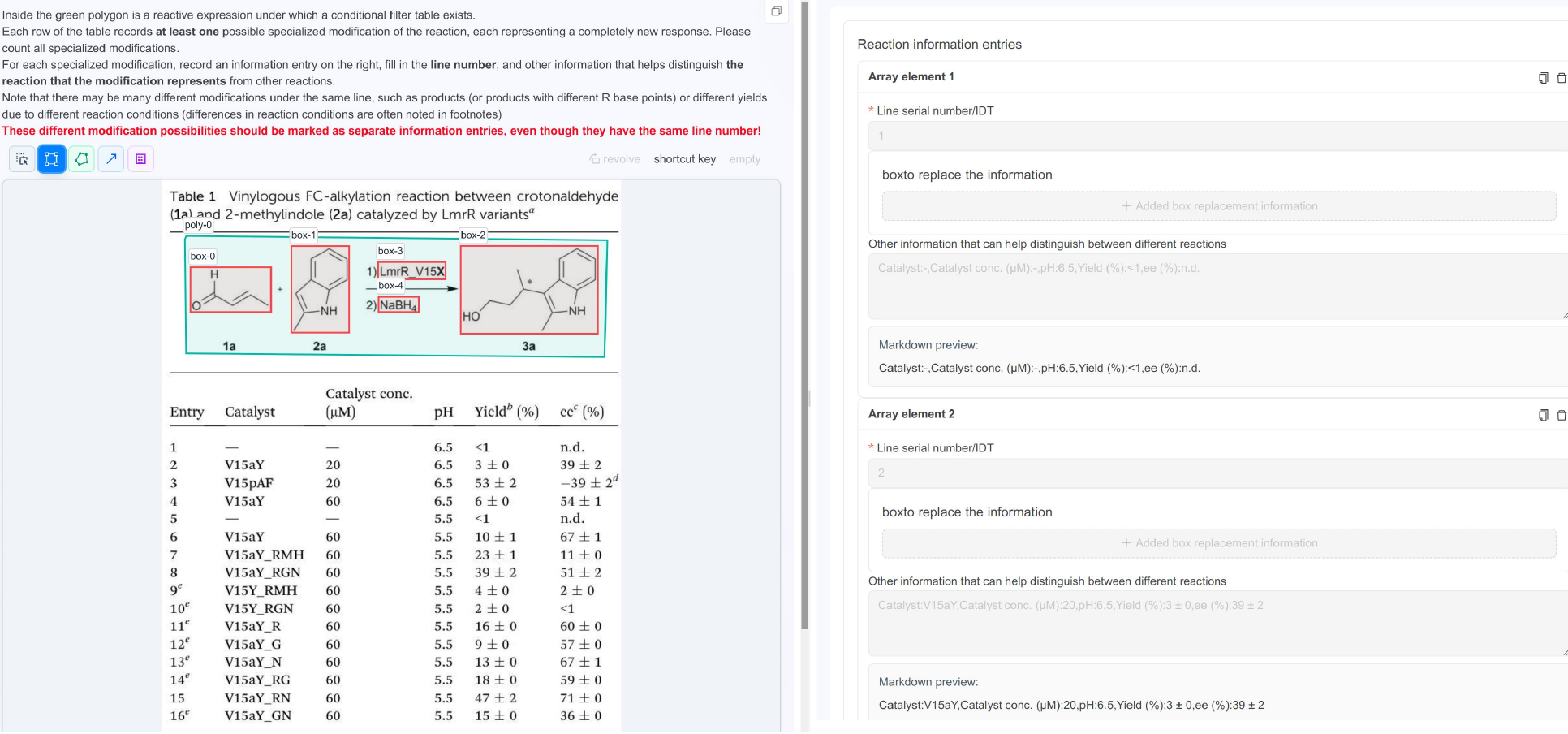}
    \caption{Screenshot for Pre-annotation of Instantiation References}
    \label{fig:appendix_c_inst_ref}
\end{figure}

\begin{figure}[!t]
    \centering
    \includegraphics[width=0.94\linewidth]{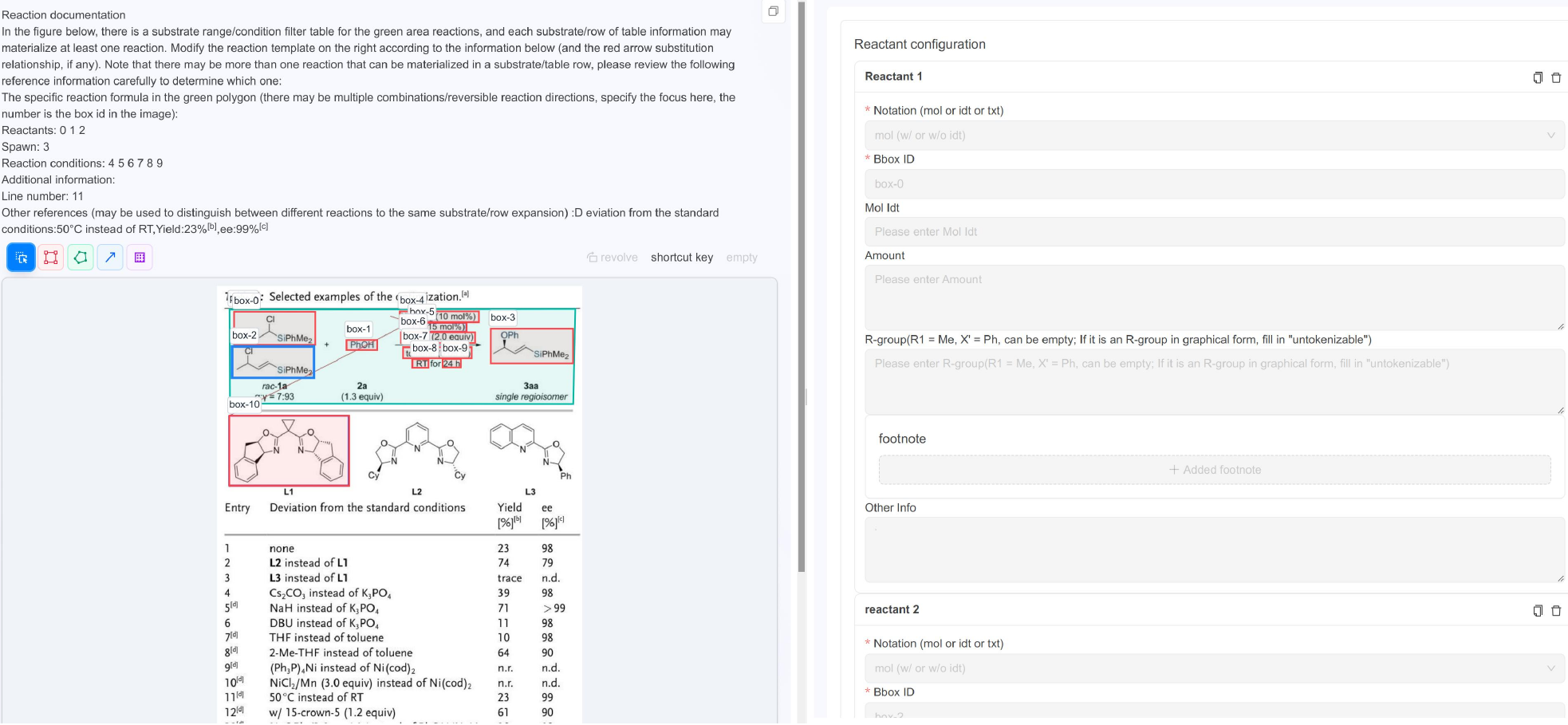}
    \caption{Screenshot for Instantiation Execution}
    \label{fig:appendix_c_rxn_inst}
\end{figure}

\begin{figure}[!t]
    \centering
    \includegraphics[width=0.94\linewidth]{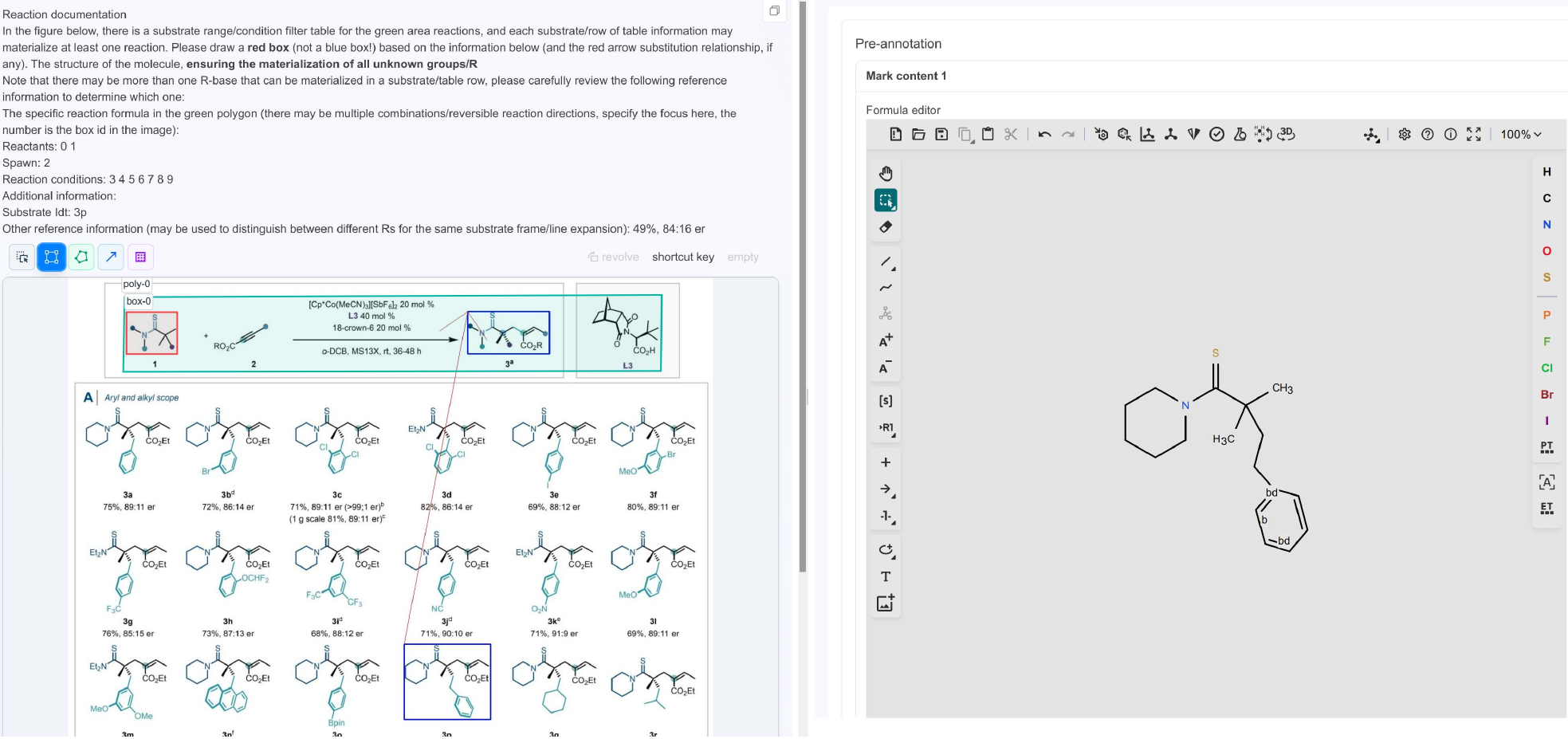}
    \caption{Screenshot for Molecular Structure (SMILES) Annotation}
    \label{fig:appendix_c_mol_edit}
\end{figure}

Each annotation step was conducted on specifically designed web-based platforms. These platforms enabled the efficient coordination of annotator team for parallel annotation, while also facilitating multiple rounds of internal quality checks and flexible data processing. Example interface screenshots for each annotation step are as follows: Figure \ref{fig:appendix_c_mol_det} shows the Molecular Structure Localization interface, Figure \ref{fig:appendix_c_rxn_abs} illustrates the Reaction Template Annotation interface, Figure \ref{fig:appendix_c_inst_ref} presents the Pre-annotation of Instantiation References interface, Figure \ref{fig:appendix_c_rxn_inst} displays the Instantiation Execution interface, and Figure \ref{fig:appendix_c_mol_edit} depicts the Molecular Structure (SMILES) Annotation interface.

\subsection{Data Annotation Team}
\label{app:C.4}
We recruited a total of 40 professional chemical annotators to participate in our chemical reaction annotation task. All annotators possess backgrounds in chemistry-related disciplines, including 9 master's degree candidates and 31 bachelor's degree holders. Their academic specialties cover multiple chemistry-related fields such as Chemistry, Applied Chemistry, Materials Chemistry, Chemical Engineering and Technology, and Organic Chemistry. Prior to formally commencing the annotation work, all annotators passed rigorous chemical knowledge assessments and comprehensive annotation protocol training. This ensures their solid foundation in chemistry and consistent adherence to annotation standards, thereby guaranteeing the professionalism and reliability of the annotated data. During the annotation process, samples were periodically cross-checked by different annotators, and disagreements were resolved through discussion and expert review. In addition, the modular design of the pipeline---which separates template annotation, instantiation, and molecular structure annotation---helps reduce ambiguity and improve consistency across different annotation stages.

All annotators were compensated under written agreements at rates that meet or exceed the prevailing wage standards for skilled scientific annotation work in their respective regions, in compliance with applicable local labor regulations.Before participating, annotators were informed in writing of the research purpose, the structure of the annotation task, the planned public release of the resulting benchmark under a CC-BY-4.0 license, and their right to withdraw at any time; all annotators provided informed consent.

\subsection{Human Performance Upper Bound}
\label{app:C.5}

To contextualize VLM results and quantify the inherent difficulty of RxnExBench, we conducted a dedicated human-performance study. Three chemistry-trained experts who did \emph{not} participate in the original annotation pipeline were asked to perform the end-to-end
extraction task on figures sampled across all three difficulty levels, following exactly the same instructions and schema given to the VLMs. Their predictions were scored against the released ground truth using the identical evaluation pipeline (Appendix~\ref{app:F}).
The averaged F1 scores are reported in Table~\ref{tab:App_C_HumanUpperBound}.

Three observations emerge:

\noindent\textbf{(i) Even trained chemists do not reach 100\%.} Overall F1 reaches 91.33\%, 85.27\%, and 81.66\% under Criteria 1, 2, and 3 respectively. The residual gap reflects the genuine ambiguity of chemical literature: footnote conventions, abbreviation
expansions, stereochemistry annotation, and tautomer choice sometimes admit more than one defensible interpretation. These numbers therefore serve as an empirical \emph{performance ceiling} for RxnExBench rather than a theoretical upper bound of 100\%.

\noindent\textbf{(ii) Human accuracy degrades monotonically with difficulty level.} Under Criterion~1, expert F1 drops from 98.98\% on Level~1 to 93.49\% on Level~2 and 87.68\% on Level~3, mirroring the same Level~1\,$>$\,Level~2\,$>$\,Level~3 ordering observed for every
VLM (Table~\ref{tab:all_result}). This empirically validates that the proposed reasoning-driven taxonomy reflects actual cognitive load rather than surface heterogeneity.

\noindent\textbf{(iii) Human accuracy also degrades monotonically with stricter criteria.} On Level~1 figures alone, F1 falls from 98.98\% (Criterion~1) to 94.92\% (Criterion~3) as the number of required-match fields grows from three to ten. This confirms that the
field-level extension from Criterion~1 to Criterion~3 introduces genuine, additive difficulty, and is not a metric artifact.

\noindent\textbf{The VLM--human gap is large and consistent.} The best VLM in our evaluation (GPT-5.5) reaches 23.16\% F1 under Criterion~1, leaving a gap of 68.17 absolute points to human performance. Under the strictest Criterion~3, the gap widens to 72.80 points
(8.86\% vs.\ 81.66\%). This sizable and consistent margin indicates that fully automated structured reaction extraction from \textit{Screen Tables} and \textit{Substrate Scopes} remains an open problem, and that RxnExBench captures challenges that current frontier VLMs
are far from solving.

\begin{table}[t]
\centering
\small
\setlength{\tabcolsep}{8pt}
\renewcommand{\arraystretch}{1.15}
\caption{Human expert performance (F1, \%) on RxnExBench, broken down by difficulty level and evaluation criterion. Numbers are averaged over three independent experts. The overall human F1 of 91.33\% under Criterion~1 serves as an empirical performance ceiling for this
benchmark.}
\label{tab:App_C_HumanUpperBound}
\begin{tabular}{lccc}
\toprule
\textbf{Subset} & \textbf{Criterion 1} & \textbf{Criterion 2} & \textbf{Criterion 3} \\
\midrule
Level 1   & 98.98\% & 96.79\% & 94.92\% \\
Level 2   & 93.49\% & 86.25\% & 81.76\% \\
Level 3   & 87.68\% & 81.09\% & 77.43\% \\
\midrule
\textbf{All} & \textbf{91.33\%} & \textbf{85.27\%} & \textbf{81.66\%} \\
\bottomrule
\end{tabular}
\end{table}

\begin{figure}[!t]
    \centering
    \includegraphics[width=0.94\linewidth]{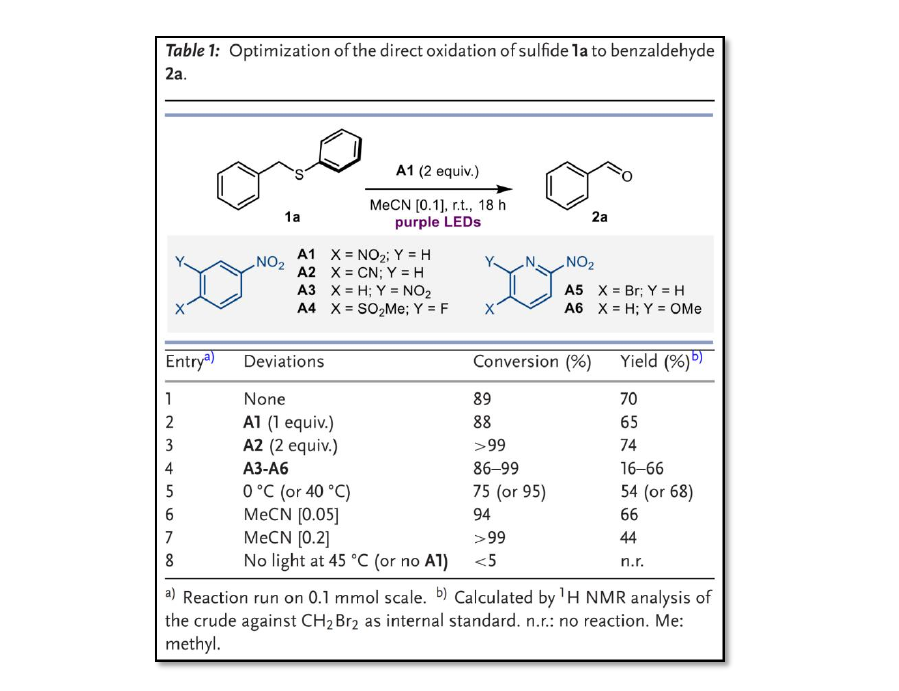}
    \caption{Example of \textit{Screen Table}. The annotation results for this \textit{Screen Table} are shown in Figures \ref{fig:GT_screentable1} and \ref{fig:GT_screentable2}.}
    \label{fig:GT_screentable_figure}
\end{figure}

\begin{figure}[!t]
    \centering
    \includegraphics[width=0.68\linewidth]{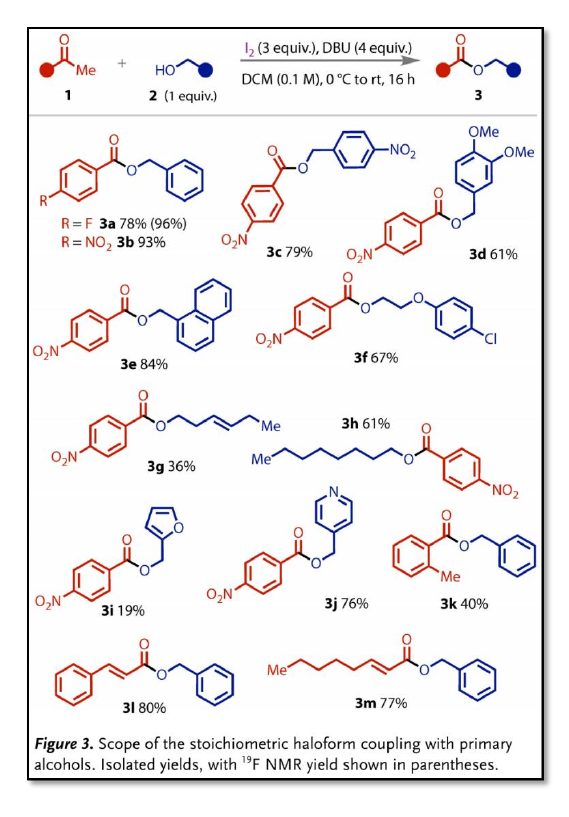}
    \caption{Example of \textit{Substrate Scope}. The annotation results for this \textit{Substrate Scope} are shown in Figures \ref{fig:GT_substrate1} and \ref{fig:GT_substrate2}.}
    \label{fig:GT_substrate_figure}
\end{figure}

\begin{table*}[h]
  \centering
  \caption{Criterion details of RxnExBench.}
  \resizebox{\linewidth}{!}{
  \begin{tabular}{p{4em} p{14em} p{18em} p{18em}}
    \toprule
    \textbf{Criterion} & \textbf{Downstream Task Category} & \textbf{Reaction Roles} & \textbf{Evaluation Description} \\
    \midrule

    1 &
      Retrosynthesis Analysis~\cite{chen2025chemical}; Main Product Prediction~\cite{coley2017prediction}; Reaction Classification~\cite{probst2022reaction} &
      reactant, product, reagent &
      Focuses on the correctness of the primary reaction roles. (SMILES or text field indicating the identity of the substance) \\
    \midrule

    2 &
      Yield Prediction~\cite{ahneman2018predicting}; Selectivity prediction~\cite{gallegos2021importance} &
      reactant, product, reagent, catalyst, solvent &
      Focuses on the correctness of reaction roles and fine-grained fields. (SMILES or text indicating the identity of the substance, and fields such as amount and yield ratio that describe quantitative reaction information.) \\
    \midrule

    3 &
      Reaction Condition Recommendation; Reaction Kinetics Modeling; HTE Screening~\cite{koscher2023autonomous,mahjour2023rapid} &
      reactant, product, reagent, catalyst, ligand, solvent, pressure, reaction\_time, reaction\_temperature, pH &
      Focuses on the correctness of reaction roles and fine-grained fields.(SMILES or text indicating the identity of the substance, and fields such as amount and yield ratio that describe quantitative reaction information.) \\
    \bottomrule
  \end{tabular}
  }
  \label{tab:App_F_Criterion}
\end{table*}

\section{Examples of RxnExBench Annotation Result}
\label{app:D}
This section presents the detailed annotation results of RxnExBench. Figure \ref{fig:GT_screentable_figure} shows an example of the \textit{Screen Table}, while Figure \ref{fig:GT_screentable1} and Figure \ref{fig:GT_screentable2} provide examples of its corresponding detailed reaction annotations.

Figure \ref{fig:GT_substrate_figure} shows an example of the  \textit{Substrate Scope}, while Figures \ref{fig:GT_substrate1} and \ref{fig:GT_substrate2} provide examples of its corresponding detailed reaction annotations.

\begin{figure}[!t]
    \centering
    \includegraphics[width=1\linewidth]{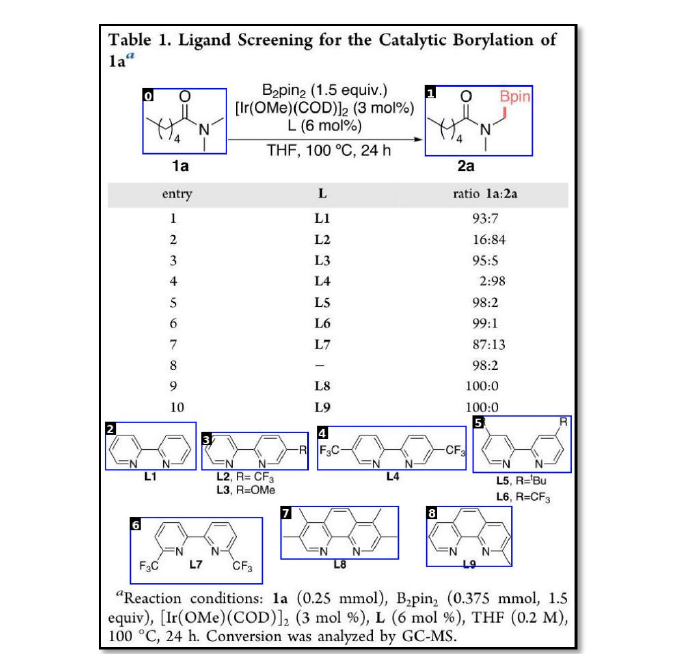}
    \caption{Example of the input image of \textsf{Oracle-MolRec-BBOX-IDX inference}. The ground-truth bounding box of each molecule is drawn on the original image as visual prompts and a unique index number is labeled at the top-left corner. The corresponding prompt for this image is shown in Figure~\ref{fig:Prompt_table_idx}.}
    \label{fig:Prompt_vp_image}
\end{figure}

\section{Inference Prompt}
\label{app:E}
This section presents all inference prompts used in \textbf{RxnExBench}.
Figures \ref{fig:Prompt_table} and \ref{fig:Prompt_scpoe} show the \textsf{standard inference} prompts for \textit{Screen Tables} and \textit{Substrate Scopes}, respectively.
Figures \ref{fig:Prompt_table_idt} and \ref{fig:Prompt_scpoe_idt} illustrate the inference prompts adopted in the \textsf{Oracle-MolRec inference} setting for \textit{Screen Tables} and \textit{Substrate Scopes}.
Figures \ref{fig:Prompt_table_idx} and \ref{fig:Prompt_scope_idx} display the inference prompts used in the \textsf{Oracle-MolRec-BBOX-IDX} setting for \textit{Screen Tables} and \textit{Substrate Scopes}.
Finally, Figures \ref{fig:Prompt_table_bbox} and \ref{fig:Prompt_scope_bbox} present the inference prompts for the \textsf{Oracle-MolRec-BBOX-CORD} setting for \textit{Screen Tables} and  \textit{Substrate Scopes}. For detailed experimental setup, please refer to Appendix~\ref{app:G}.

\begin{table*}[h]
    \centering
    \scriptsize
    \renewcommand{\arraystretch}{1.3}
  \caption{Recall scores of VLMs on RxnExBench. Results are shown for the full dataset and three difficulty levels under three evaluation criteria. \textbf{C} refers to \textbf{Criterion}. \textbf{Boldfaced} numbers denote the best performance, while \underline{underlined} numbers represent the runner-up.}
    \newcolumntype{C}{>{\centering\arraybackslash}X}
      \begin{tabularx}{\textwidth}{p{8.5em} | *{3}{C} | *{3}{C} | *{3}{C} | *{3}{C}}
      \toprule
      \multirow{2}{*}{\textbf{Models}}
      & \multicolumn{3}{c|}{\textbf{All Levels (266 Figures)}}
      & \multicolumn{3}{c|}{\textbf{Level 1 (60 Figures)}}
      & \multicolumn{3}{c|}{\textbf{Level 2 (90 Figures)}}
      & \multicolumn{3}{c}{\textbf{Level 3 (116 Figures)}} \\
      & C1 & C2 & C3
      & C1 & C2 & C3
      & C1 & C2 & C3
      & C1 & C2 & C3 \\
      \midrule
      \rowcolors{1}{gray!10}{white}
      Claude Opus 4.6 & 2.50\% & 1.07\% & 0.93\% & 5.58\% & 4.23\% & 4.23\% & 2.38\% & 1.19\% & 0.73\% & 1.59\% & 0.00\% & 0.00\% \\
      Gemini 3.1 Pro & 18.99\% & \textbf{9.34\%} & \textbf{8.64\%} & \textbf{39.26\%} & \textbf{19.97\%} & \textbf{19.29\%} & 15.12\% & \underline{9.99\%} & \textbf{8.07\%} & \underline{14.87\%} & \textbf{5.63\%} & \textbf{5.63\%} \\
      GPT-5.5 & \textbf{22.97\%} & \underline{9.09\%} & \underline{6.76\%} & \underline{33.67\%} & \underline{12.86\%} & \underline{12.69\%} & \textbf{28.14\%} & \textbf{13.93\%} & \underline{6.42\%} & \textbf{16.62\%} & \underline{5.10\%} & \underline{5.10\%} \\
      GPT-4o & 0.08\% & 0.00\% & 0.00\% & 0.00\% & 0.00\% & 0.00\% & 0.00\% & 0.00\% & 0.00\% & 0.16\% & 0.00\% & 0.00\% \\
      Doubao Seed 2.0 Pro & \underline{19.30\%} & 6.42\% & 5.22\% & 33.50\% & 11.34\% & 10.66\% & \underline{25.39\%} & 7.06\% & 4.49\% & 11.31\% & 4.51\% & 3.93\% \\
      GLM-5V Turbo & 4.74\% & 1.80\% & 1.68\% & 15.40\% & 6.77\% & 6.26\% & 2.75\% & 1.28\% & 1.19\% & 2.55\% & 0.53\% & 0.53\% \\
      Intern-S1 & 2.83\% & 0.06\% & 0.06\% & 4.57\% & 0.00\% & 0.00\% & 6.78\% & 0.18\% & 0.18\% & 0.00\% & 0.00\% & 0.00\% \\
      Llama 4 Maverick & 1.99\% & 0.76\% & 0.76\% & 4.57\% & 2.37\% & 2.37\% & 2.66\% & 0.09\% & 0.09\% & 0.80\% & 0.64\% & 0.64\% \\
      Qwen3-VL-235B & 1.60\% & 0.25\% & 0.06\% & 5.58\% & 1.52\% & 0.34\% & 2.20\% & 0.00\% & 0.00\% & 0.00\% & 0.00\% & 0.00\% \\
      \bottomrule
    \end{tabularx}
  \label{tab:Appendix_G_recall}
  \end{table*}
\begin{table*}[h]
    \centering
    \scriptsize
    \renewcommand{\arraystretch}{1.3}
  \caption{Precision scores of VLMs on RxnExBench. Results are shown for the full dataset and three difficulty levels under three evaluation criteria. \textbf{C} refers to \textbf{Criterion}. \textbf{Boldfaced} numbers denote the best performance, while \underline{underlined} numbers represent the runner-up.}
    \newcolumntype{C}{>{\centering\arraybackslash}X}
      \begin{tabularx}{\textwidth}{p{8.5em} | *{3}{C} | *{3}{C} | *{3}{C} | *{3}{C}}
      \toprule
      \multirow{2}{*}{\textbf{Models}}
      & \multicolumn{3}{c|}{\textbf{All Levels (266 Figures)}}
      & \multicolumn{3}{c|}{\textbf{Level 1 (60 Figures)}}
      & \multicolumn{3}{c|}{\textbf{Level 2 (90 Figures)}}
      & \multicolumn{3}{c}{\textbf{Level 3 (116 Figures)}} \\
      & C1 & C2 & C3
      & C1 & C2 & C3
      & C1 & C2 & C3
      & C1 & C2 & C3 \\
      \midrule
      \rowcolors{1}{gray!10}{white}
      Claude Opus 4.6 & 2.55\% & 1.09\% & 0.95\% & 5.61\% & 4.25\% & 4.25\% & 2.42\% & 1.21\% & 0.74\% & 1.64\% & 0.00\% & 0.00\% \\
      Gemini 3.1 Pro & 19.96\% & \textbf{9.82\%} & \textbf{9.08\%} & \textbf{41.65\%} & \textbf{21.18\%} & \textbf{20.47\%} & 15.82\% & \underline{10.45\%} & \textbf{8.44\%} & \underline{15.63\%} & \textbf{5.92\%} & \textbf{5.92\%} \\
      GPT-5.5 & \textbf{23.34\%} & \underline{9.23\%} & \underline{6.87\%} & 33.33\% & \underline{12.73\%} & \underline{12.56\%} & \textbf{28.11\%} & \textbf{13.92\%} & \underline{6.41\%} & \textbf{17.20\%} & \underline{5.27\%} & \underline{5.27\%} \\
      GPT-4o & 0.09\% & 0.00\% & 0.00\% & 0.00\% & 0.00\% & 0.00\% & 0.00\% & 0.00\% & 0.00\% & 0.17\% & 0.00\% & 0.00\% \\
      Doubao Seed 2.0 Pro & \underline{20.34\%} & 6.77\% & 5.50\% & \underline{33.62\%} & 11.38\% & 10.70\% & \underline{26.01\%} & 7.23\% & 4.60\% & 12.33\% & 4.92\% & 4.28\% \\
      GLM-5V Turbo & 5.47\% & 2.07\% & 1.94\% & 15.56\% & 6.84\% & 6.32\% & 2.79\% & 1.30\% & 1.21\% & 3.36\% & 0.70\% & 0.70\% \\
      Intern-S1 & 2.93\% & 0.06\% & 0.06\% & 4.62\% & 0.00\% & 0.00\% & 7.01\% & 0.19\% & 0.19\% & 0.00\% & 0.00\% & 0.00\% \\
      Llama 4 Maverick & 2.65\% & 1.01\% & 1.01\% & 4.60\% & 2.39\% & 2.39\% & 3.03\% & 0.10\% & 0.10\% & 1.32\% & 1.06\% & 1.06\% \\
      Qwen3-VL-235B & 1.76\% & 0.28\% & 0.06\% & 5.65\% & 1.54\% & 0.34\% & 2.27\% & 0.00\% & 0.00\% & 0.00\% & 0.00\% & 0.00\% \\
      \bottomrule
    \end{tabularx}
  \label{tab:Appendix_G_precision}
  \end{table*}

\begin{table*}[h]
    \centering
    \scriptsize
    \renewcommand{\arraystretch}{1.3}
  \caption{The Recall scores of each VLM under the four inference settings: \textsf{Standard Inference}, \textsf{Oracle-MolRec-IDT}, \textsf{Oracle-MolRec-BBOX-CORD}, and \textsf{Oracle-MolRec-BBOX-IDX}, on the \textbf{RxnExBench-MolIDT} subset. \textbf{C} refers to \textbf{Criterion}. \textbf{Boldfaced} numbers denote the best performance, while \underline{underlined} numbers represent the runner-up.}
    \newcolumntype{C}{>{\centering\arraybackslash}X}
    \begin{tabularx}{\textwidth}{p{8em} | *{3}{C} | *{3}{C} | *{3}{C} | *{3}{C}}
      \toprule
      \multirow{2}{*}{\textbf{Models}}
      & \multicolumn{3}{c|}{\textbf{Standard Inference}}
      & \multicolumn{3}{c|}{\textbf{Oracle-MolRec-IDT}}
      & \multicolumn{3}{c|}{\textbf{Oracle-MolRec-BBOX-CORD}}
      & \multicolumn{3}{c}{\textbf{Oracle-MolRec-BBOX-IDX}} \\
      & C1 & C2 & C3
      & C1 & C2 & C3
      & C1 & C2 & C3
      & C1 & C2 & C3 \\
      \midrule
      \rowcolors{1}{gray!10}{white}
      Claude Opus 4.6 & 2.50\% & 1.07\% & 0.93\% & 29.63\% & \underline{16.96\%} & 14.58\% & 28.83\% & 14.77\% & 12.31\% & 29.98\% & \underline{15.13\%} & \underline{12.95\%} \\
      Gemini 3.1 Pro & 18.99\% & \textbf{9.34\%} & \textbf{8.64\%} & 35.15\% & \textbf{17.47\%} & \textbf{15.21\%} & 31.85\% & \textbf{16.32\%} & \underline{13.98\%} & 34.99\% & \textbf{19.18\%} & \textbf{16.16\%} \\
      GPT-5.5 & \textbf{22.97\%} & \underline{9.09\%} & \underline{6.76\%} & \textbf{41.78\%} & 16.48\% & \underline{14.97\%} & \textbf{45.04\%} & \underline{16.12\%} & \textbf{14.26\%} & \textbf{45.75\%} & 14.97\% & 12.67\% \\
      GPT-4o & 0.08\% & 0.00\% & 0.00\% & 24.15\% & 5.08\% & 4.37\% & 22.16\% & 5.36\% & 4.25\% & 21.37\% & 4.01\% & 2.94\% \\
      Doubao Seed 2.0 Pro & \underline{19.30\%} & 6.42\% & 5.22\% & \underline{35.31\%} & 11.48\% & 10.01\% & \underline{35.19\%} & 11.32\% & 10.44\% & \underline{36.30\%} & 10.56\% & 9.93\% \\
      GLM-5V Turbo & 4.74\% & 1.80\% & 1.68\% & 31.53\% & 11.83\% & 10.96\% & 30.30\% & 10.64\% & 9.13\% & 33.36\% & 9.21\% & 7.86\% \\
      Intern-S1 & 2.83\% & 0.06\% & 0.06\% & 18.71\% & 6.67\% & 5.28\% & 20.77\% & 6.55\% & 6.04\% & 19.86\% & 5.68\% & 4.96\% \\
      Llama 4 Maverick & 1.99\% & 0.76\% & 0.76\% & 17.51\% & 7.82\% & 7.11\% & 14.42\% & 5.36\% & 4.57\% & 19.06\% & 8.18\% & 6.63\% \\
      Qwen3-VL-235B & 1.60\% & 0.25\% & 0.06\% & 25.54\% & 6.87\% & 5.64\% & 23.23\% & 6.16\% & 5.08\% & 25.34\% & 7.70\% & 6.24\% \\
      \bottomrule
    \end{tabularx}
  \label{tab:subset_recall}
  \end{table*}

% Table generated by Excel2LaTeX from sheet 'Sheet3'
\begin{table}[h]
  \small
  \centering
  \caption{Results of Mermaid on RxnExBench}
    \begin{tabular}{cccc}
    \toprule
    \textbf{Criterion} & \multicolumn{1}{p{4em}}{\textbf{Recall}} & \multicolumn{1}{p{4em}}{\textbf{Precision}} & \multicolumn{1}{p{4em}}{\textbf{F1-score}} \\
    \midrule
    \textbf{Criterion1} & 3.99\% & 4.68\% & 4.31\% \\
    \textbf{Criterion2} & 0.59\% & 0.69\% & 0.64\% \\
    \textbf{Criterion3} & 0.59\% & 0.69\% & 0.64\% \\
    \bottomrule
    \end{tabular}%
  \label{tab:mermaid_res}%
\end{table}%

\begin{table}[h]
  \centering
  \small
  \caption{Results of ChemEAGLE on RxnExBench}
    \begin{tabular}{cccc}
    \toprule
    \textbf{Criterion} & \multicolumn{1}{p{4em}}{\textbf{Recall}} & \multicolumn{1}{p{4em}}{\textbf{Precision}} & \multicolumn{1}{p{4em}}{\textbf{F1-score}} \\
    \midrule
    \textbf{Criterion1} & 0.31\% & 0.43\% & 0.36\% \\
    \textbf{Criterion2} & 0.00\% & 0.00\% & 0.00\% \\
    \textbf{Criterion3} & 0.00\% & 0.00\% & 0.00\% \\
    \bottomrule
    \end{tabular}%
  \label{tab:chemeagle_res}%
\end{table}%
\begin{table*}[h]
    \centering
    \scriptsize
    \renewcommand{\arraystretch}{1.3}
  \caption{The Precision scores of each VLM under the four inference settings: \textsf{Standard Inference}, \textsf{Oracle-MolRec-IDT}, \textsf{Oracle-MolRec-BBOX-CORD}, and \textsf{Oracle-MolRec-BBOX-IDX}, on the \textbf{RxnExBench-MolIDT} subset. \textbf{C} refers to \textbf{Criterion}. \textbf{Boldfaced} numbers denote the best performance, while \underline{underlined} numbers represent the runner-up.}
    \newcolumntype{C}{>{\centering\arraybackslash}X}
    \begin{tabularx}{\textwidth}{p{8em} | *{3}{C} | *{3}{C} | *{3}{C} | *{3}{C}}
      \toprule
      \multirow{2}{*}{\textbf{Models}}
      & \multicolumn{3}{c|}{\textbf{Standard Inference}}
      & \multicolumn{3}{c|}{\textbf{Oracle-MolRec-IDT}}
      & \multicolumn{3}{c|}{\textbf{Oracle-MolRec-BBOX-CORD}}
      & \multicolumn{3}{c}{\textbf{Oracle-MolRec-BBOX-IDX}} \\
      & C1 & C2 & C3
      & C1 & C2 & C3
      & C1 & C2 & C3
      & C1 & C2 & C3 \\
      \midrule
      \rowcolors{1}{gray!10}{white}
      Claude Opus 4.6 & 2.55\% & 1.09\% & 0.95\% & 30.47\% & \underline{17.44\%} & 14.99\% & 29.50\% & 15.12\% & 12.60\% & 30.80\% & \underline{15.54\%} & \underline{13.30\%} \\
      Gemini 3.1 Pro & 19.96\% & \textbf{9.82\%} & \textbf{9.08\%} & \underline{38.39\%} & \textbf{19.09\%} & \textbf{16.62\%} & 35.98\% & \textbf{18.44\%} & \textbf{15.79\%} & \underline{38.83\%} & \textbf{21.29\%} & \textbf{17.94\%} \\
      GPT-5.5 & \textbf{23.34\%} & \underline{9.23\%} & \underline{6.87\%} & \textbf{43.60\%} & 17.20\% & \underline{15.62\%} & \textbf{45.89\%} & \underline{16.43\%} & \underline{14.53\%} & \textbf{46.30\%} & 15.15\% & 12.82\% \\
      GPT-4o & 0.09\% & 0.00\% & 0.00\% & 26.46\% & 5.57\% & 4.79\% & 24.37\% & 5.90\% & 4.67\% & 23.41\% & 4.40\% & 3.22\% \\
      Doubao Seed 2.0 Pro & \underline{20.34\%} & 6.77\% & 5.50\% & 38.27\% & 12.44\% & 10.85\% & \underline{37.88\%} & 12.18\% & 11.24\% & 38.36\% & 11.16\% & 10.49\% \\
      GLM-5V Turbo & 5.47\% & 2.07\% & 1.94\% & 32.38\% & 12.15\% & 11.26\% & 31.08\% & 10.92\% & 9.37\% & 34.30\% & 9.47\% & 8.08\% \\
      Intern-S1 & 2.93\% & 0.06\% & 0.06\% & 23.35\% & 8.33\% & 6.59\% & 22.00\% & 6.94\% & 6.39\% & 21.84\% & 6.25\% & 5.46\% \\
      Llama 4 Maverick & 2.65\% & 1.01\% & 1.01\% & 24.24\% & 10.83\% & 9.84\% & 20.10\% & 7.48\% & 6.37\% & 27.03\% & 11.60\% & 9.40\% \\
      Qwen3-VL-235B & 1.76\% & 0.28\% & 0.06\% & 28.41\% & 7.64\% & 6.27\% & 25.98\% & 6.88\% & 5.68\% & 28.84\% & 8.77\% & 7.10\% \\
      \bottomrule
    \end{tabularx}
  \label{tab:subset_precision}
  \end{table*}

\begin{table*}[h]
    \centering
    \scriptsize
    \renewcommand{\arraystretch}{1.3}
  \caption{The F1 scores of each VLM under the four inference settings: \textsf{Standard Inference}, \textsf{Oracle-MolRec-IDT}, \textsf{Oracle-MolRec-BBOX-CORD}, and \textsf{Oracle-MolRec-BBOX-IDX}, on the \textbf{RxnExBench-MolIDT} subset. \textbf{C} refers to \textbf{Criterion}. \textbf{Boldfaced} numbers denote the best performance, while \underline{underlined} numbers represent the runner-up.}
    \newcolumntype{C}{>{\centering\arraybackslash}X}
    \begin{tabularx}{\textwidth}{p{8em} | *{3}{C} | *{3}{C} | *{3}{C} | *{3}{C}}
      \toprule
      \multirow{2}{*}{\textbf{Models}}
      & \multicolumn{3}{c|}{\textbf{Standard Inference}}
      & \multicolumn{3}{c|}{\textbf{Oracle-MolRec-IDT}}
      & \multicolumn{3}{c|}{\textbf{Oracle-MolRec-BBOX-CORD}}
      & \multicolumn{3}{c}{\textbf{Oracle-MolRec-BBOX-IDX}} \\
      & C1 & C2 & C3
      & C1 & C2 & C3
      & C1 & C2 & C3
      & C1 & C2 & C3 \\
      \midrule
      \rowcolors{1}{gray!10}{white}
      Claude Opus 4.6 & 2.52\% & 1.08\% & 0.94\% & 30.04\% & \underline{17.20\%} & 14.78\% & 29.16\% & 14.94\% & 12.45\% & 30.39\% & \underline{15.34\%} & \underline{13.12\%} \\
      Gemini 3.1 Pro & 19.47\% & \textbf{9.57\%} & \textbf{8.86\%} & 36.70\% & \textbf{18.25\%} & \textbf{15.88\%} & 33.79\% & \textbf{17.32\%} & \textbf{14.83\%} & 36.81\% & \textbf{20.18\%} & \textbf{17.00\%} \\
      GPT-5.5 & \textbf{23.16\%} & \underline{9.16\%} & \underline{6.81\%} & \textbf{42.67\%} & 16.83\% & \underline{15.29\%} & \textbf{45.46\%} & \underline{16.28\%} & \underline{14.39\%} & \textbf{46.02\%} & 15.06\% & 12.74\% \\
      GPT-4o & 0.09\% & 0.00\% & 0.00\% & 25.25\% & 5.32\% & 4.57\% & 23.21\% & 5.62\% & 4.45\% & 22.34\% & 4.19\% & 3.07\% \\
      Doubao Seed 2.0 Pro & \underline{19.81\%} & 6.59\% & 5.35\% & \underline{36.73\%} & 11.94\% & 10.41\% & \underline{36.48\%} & 11.74\% & 10.83\% & \underline{37.30\%} & 10.85\% & 10.20\% \\
      GLM-5V Turbo & 5.08\% & 1.92\% & 1.80\% & 31.95\% & 11.99\% & 11.11\% & 30.69\% & 10.78\% & 9.25\% & 33.82\% & 9.34\% & 7.97\% \\
      Intern-S1 & 2.88\% & 0.06\% & 0.06\% & 20.77\% & 7.41\% & 5.87\% & 21.37\% & 6.74\% & 6.21\% & 20.80\% & 5.95\% & 5.20\% \\
      Llama 4 Maverick & 2.27\% & 0.86\% & 0.86\% & 20.34\% & 9.08\% & 8.25\% & 16.79\% & 6.24\% & 5.32\% & 22.36\% & 9.59\% & 7.78\% \\
      Qwen3-VL-235B & 1.68\% & 0.26\% & 0.06\% & 26.90\% & 7.24\% & 5.94\% & 24.53\% & 6.50\% & 5.37\% & 26.98\% & 8.20\% & 6.64\% \\
      \bottomrule
    \end{tabularx}
  \label{tab:subset_f1}
  \end{table*}

\section{Criterion and evaluation details of RxnExBench}
\label{app:F}
As shown in Table \ref{tab:App_F_Criterion}, we design a three-level evaluation criterion based on the requirements of downstream AI-for-Chemistry tasks. \textbf{Criterion~1}, targeting tasks such as single-step retrosynthesis, checks only the reactant, product, and reagent fields. \textbf{Criterion~2}, used for tasks such as yield prediction, evaluates all fields in Criterion~1 and additionally checks catalyst and solvent information. \textbf{Criterion~3} is the most stringent: designed for tasks such as reaction-condition recommendation, it requires verifying a total of ten fields. Under all three criteria, a prediction is counted as a true positive (TP) only when it exactly matches the ground-truth annotation across all required fields.

For SMILES fields, we use \texttt{RDKit} to canonicalize all molecular representations before comparison. Two molecules are considered equivalent only if their canonicalized molecular graphs are identical, implemented by computing Tanimoto similarity on RDKit fingerprints with a strict threshold of 1.0. For other string-based fields (e.g., textual descriptions and reaction conditions), we apply a normalized string-matching procedure to improve robustness against superficial formatting variations. Specifically, strings are lowercased and normalized by removing redundant whitespace and standardizing common symbols and delimiters (e.g., commas, semicolons, and temperature/unit formatting). Matching is then performed on the normalized strings while preserving the original semantic content. This strategy tolerates minor formatting inconsistencies without relaxing the underlying field semantics.

The Tanimoto similarity is used to measure the similarity between the fingerprints of two molecules, and is defined as
\[
T(A,B) = \frac{|A \cap B|}{|A| + |B| - |A \cap B|}
\]
where \(A\) and \(B\) are the binary fingerprints of the two molecules, and \(|A \cap B|\) denotes the number of bits that are 1 in both fingerprints.

Tanimoto @1.0 indicates that two molecules are considered chemically equivalent only if their fingerprints match exactly, i.e., the Tanimoto similarity equals 1.0.

\section{Detailed experimental results}
\label{app:G}

\begin{figure*}[b]
    \centering
    \includegraphics[width=1\linewidth]{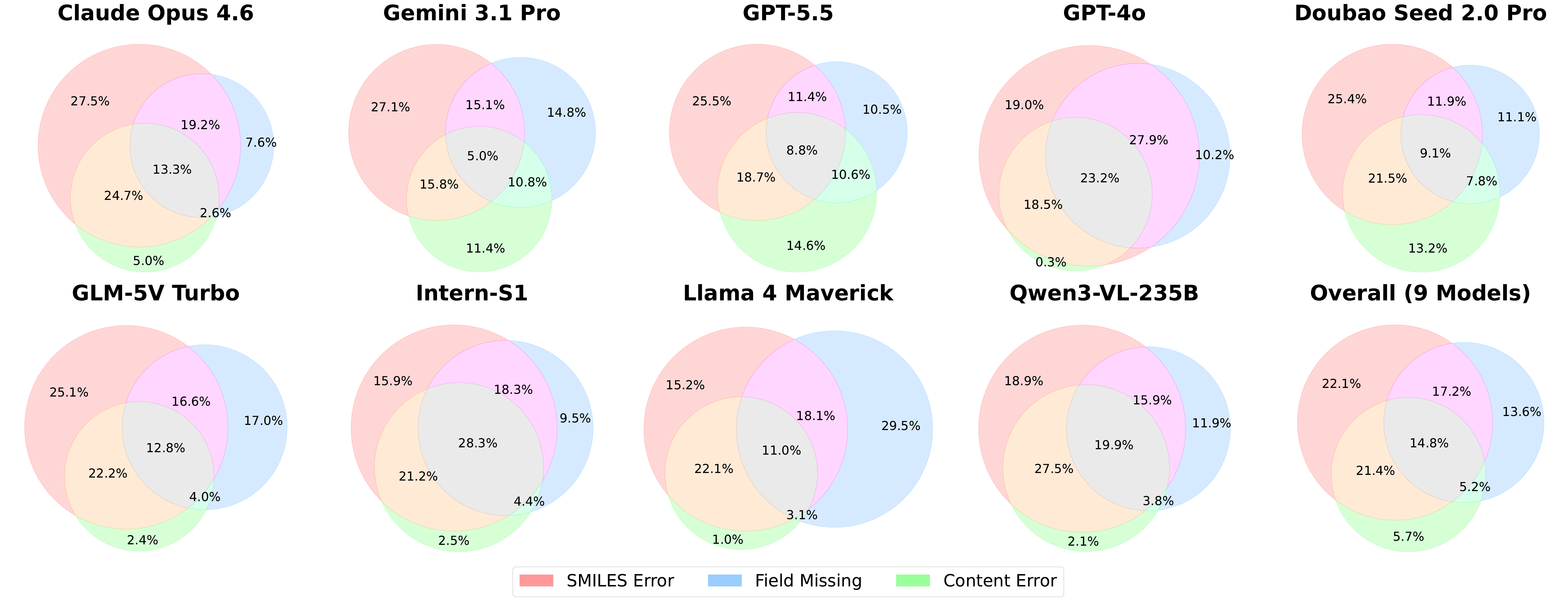}
    \caption{Venn diagrams showing error type distribution for each model. The diagrams illustrate the distribution and overlap of three error types: SMILES Error, Field Missing, and Content Error. Percentages indicate the proportion of each error category among all errors.}
    \label{fig:Appendix_G_all_venn_diagrams}
\end{figure*}

\begin{figure*}[!t]
    \centering
    \includegraphics[width=0.9\linewidth]{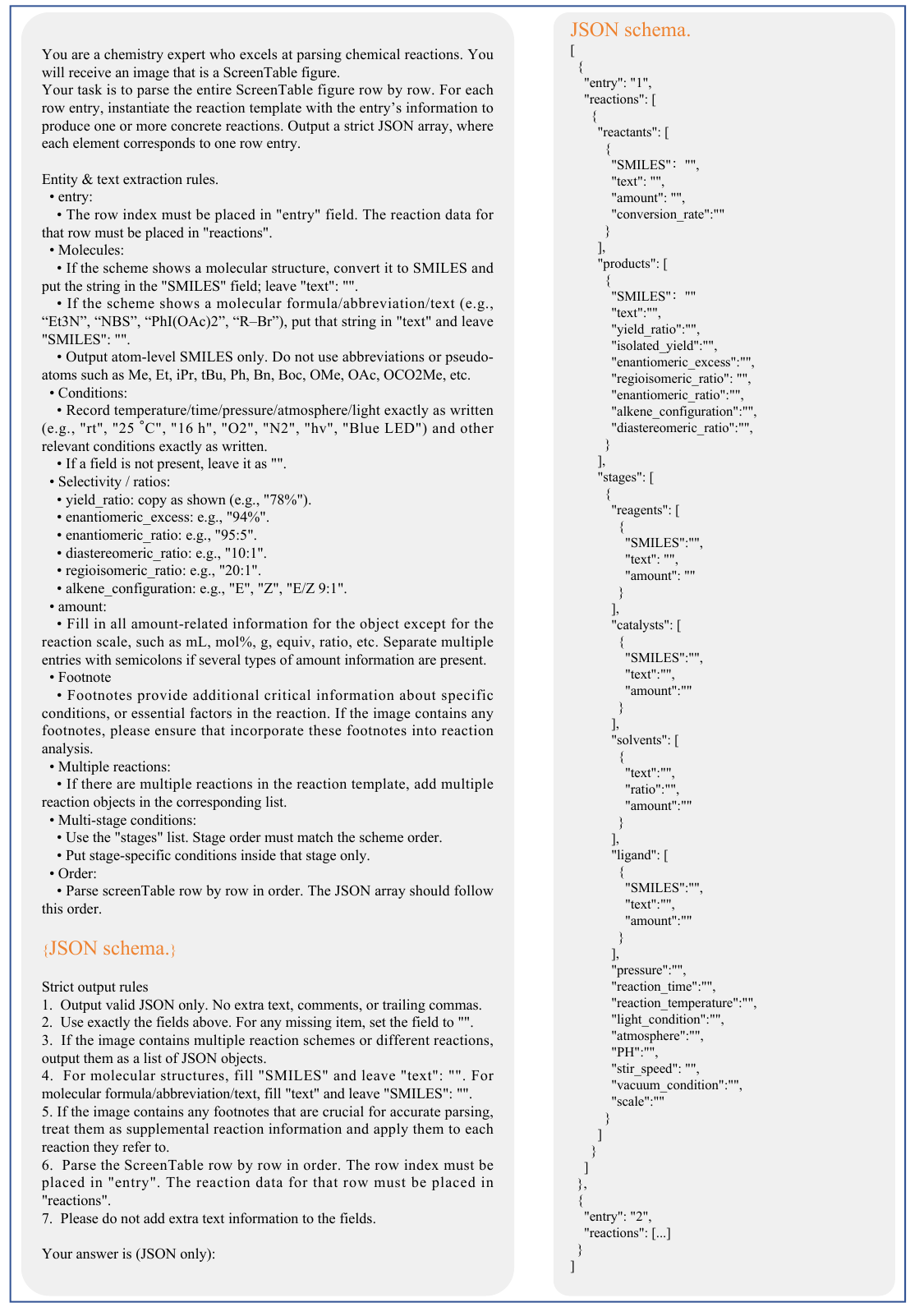}
     \caption{\textsf{Standard inference} prompt of \textit{Screen Table}}
    \label{fig:Prompt_table}
\end{figure*}

\begin{figure*}[!t]
    \centering
    \includegraphics[width=0.9\linewidth]{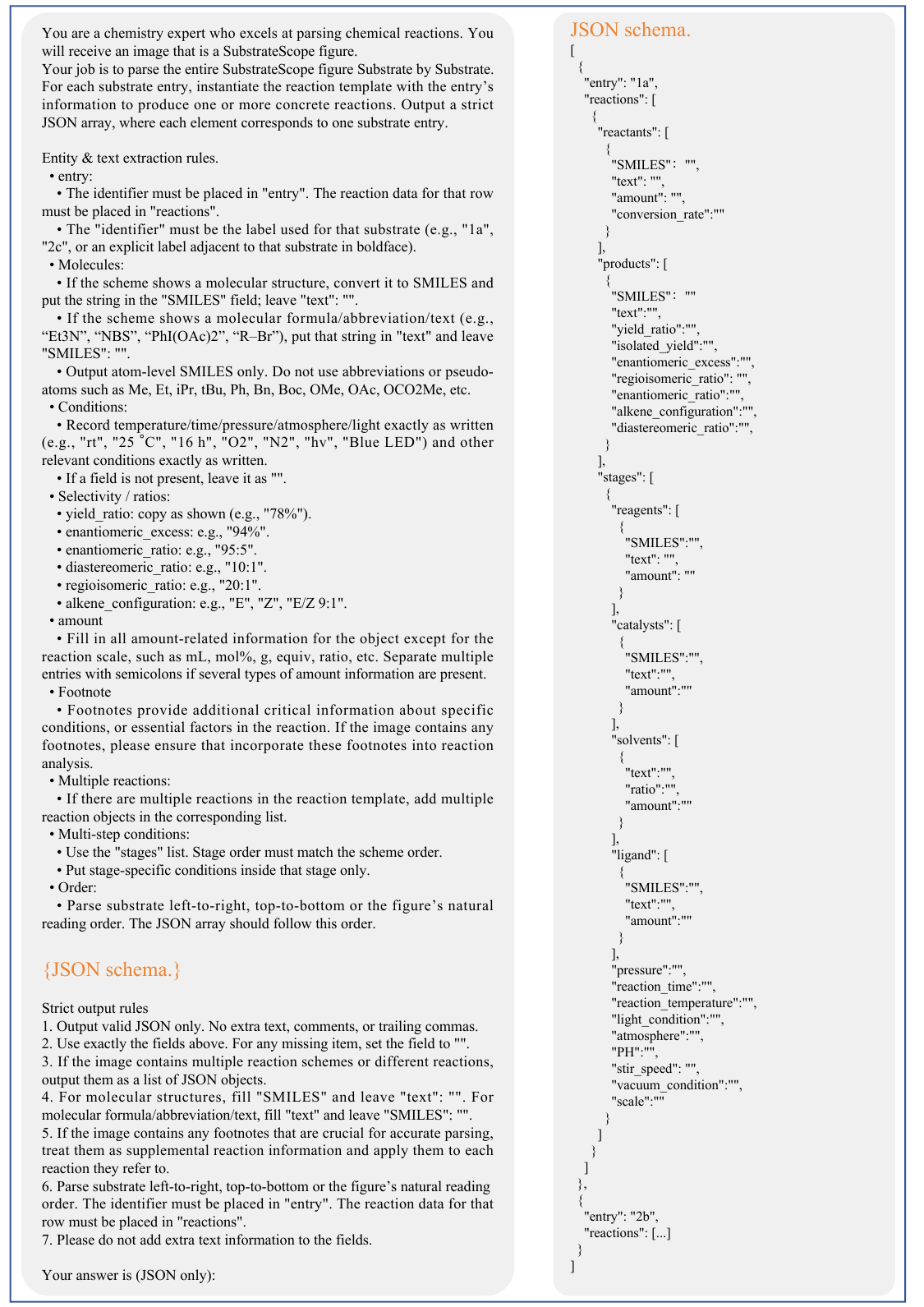}
     \caption{\textsf{Standard inference} prompt of \textit{Substrate Scope}}
    \label{fig:Prompt_scpoe}
\end{figure*}

\begin{figure*}[!t]
    \centering
    \includegraphics[width=0.9\linewidth]{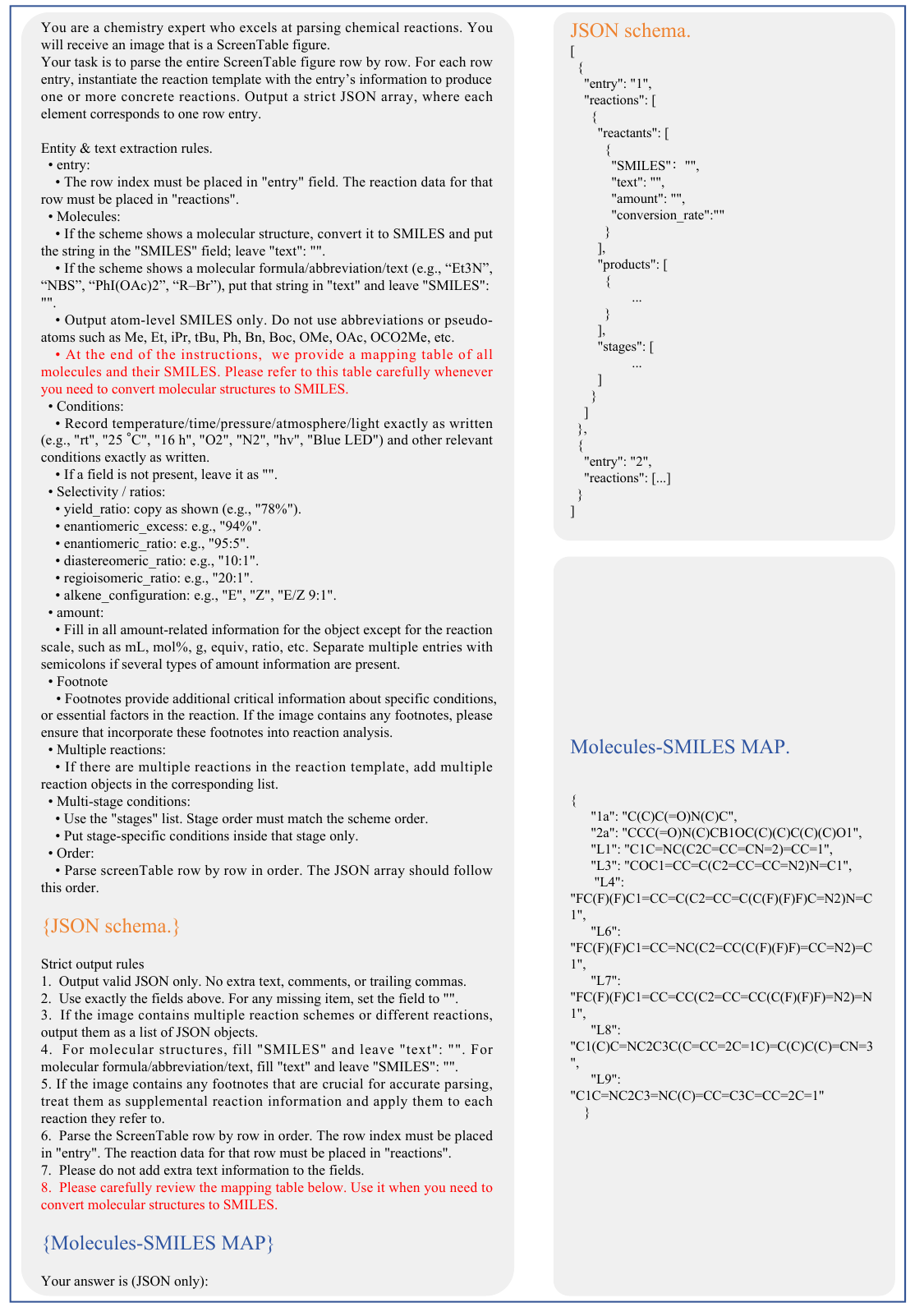}
    \caption{\textsf{Oracle-MolRec-IDT inference} prompt of \textit{Screen Table}}
    \label{fig:Prompt_table_idt}
\end{figure*}

\begin{figure*}[!t]
    \centering
    \includegraphics[width=0.9\linewidth]{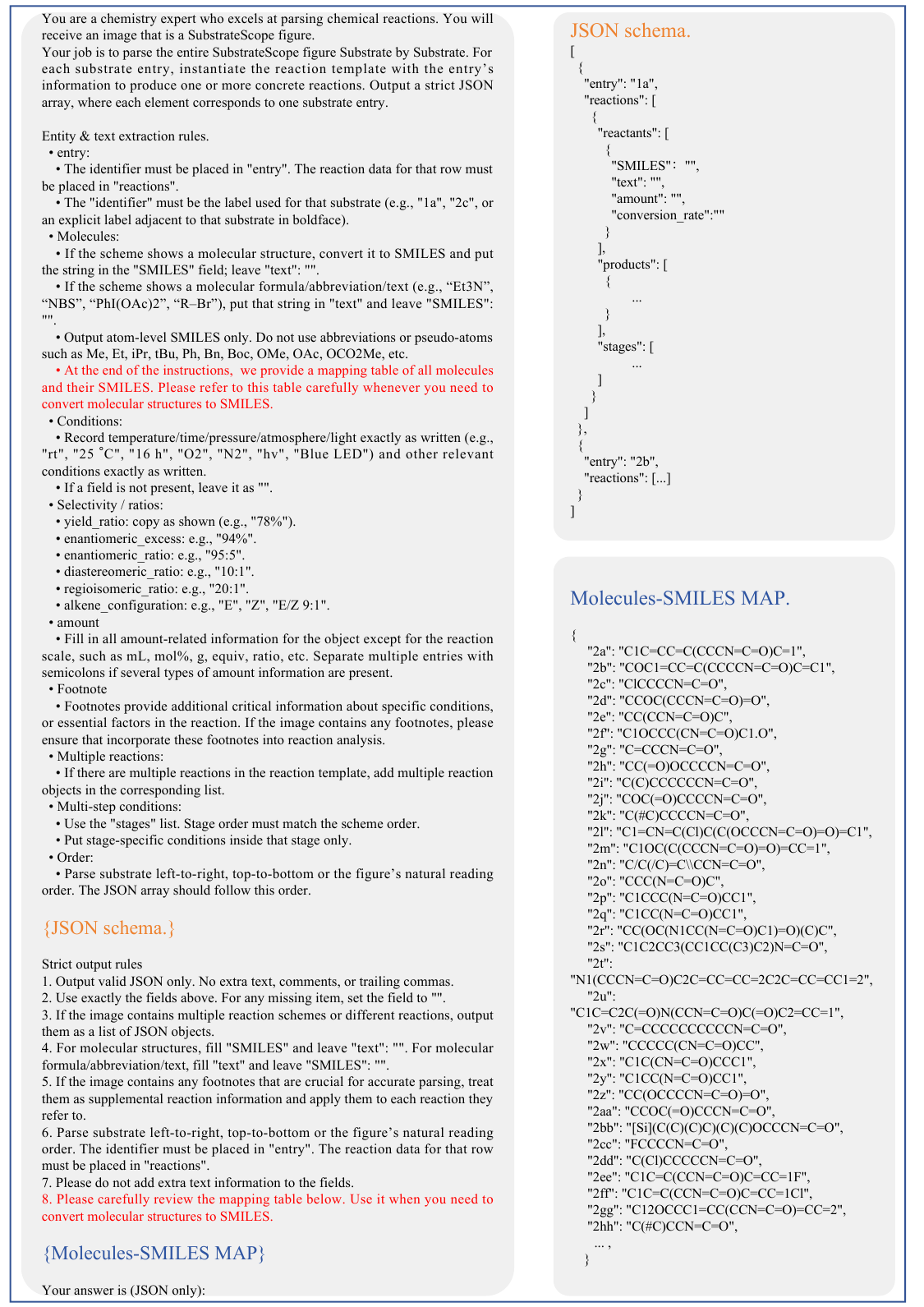}
    \caption{\textsf{Oracle-MolRec-IDT inference} prompt of \textit{Substrate Scope}}
    \label{fig:Prompt_scpoe_idt}
\end{figure*}

\begin{figure*}[!t]
    \centering
    \includegraphics[width=0.9\linewidth]{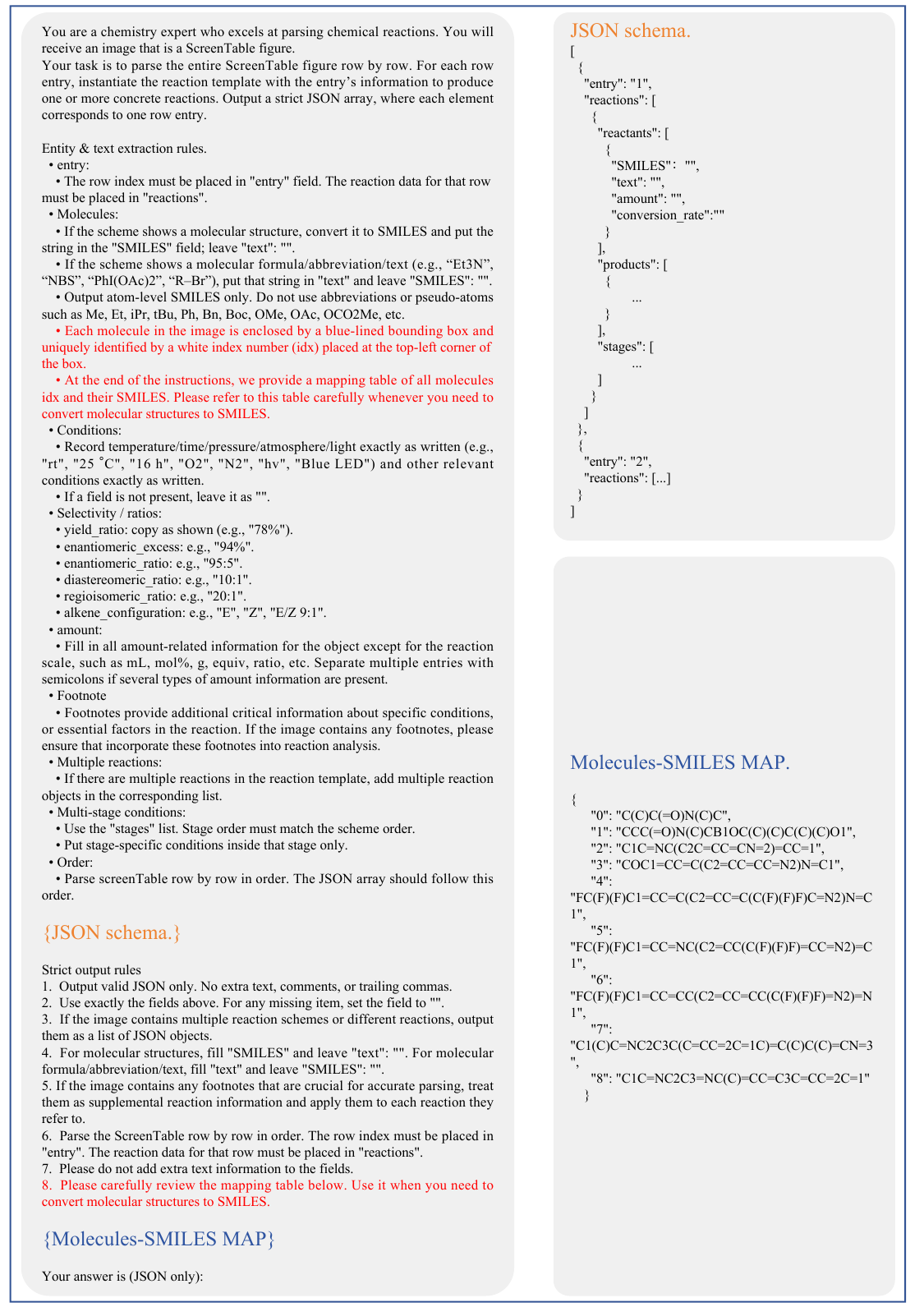}
    \caption{\textsf{Oracle-MolRec-BBOX-IDX inference} prompt of \textit{Screen Table}}
    \label{fig:Prompt_table_idx}
\end{figure*}

\begin{figure*}[!t]
    \centering
    \includegraphics[width=0.9\linewidth]{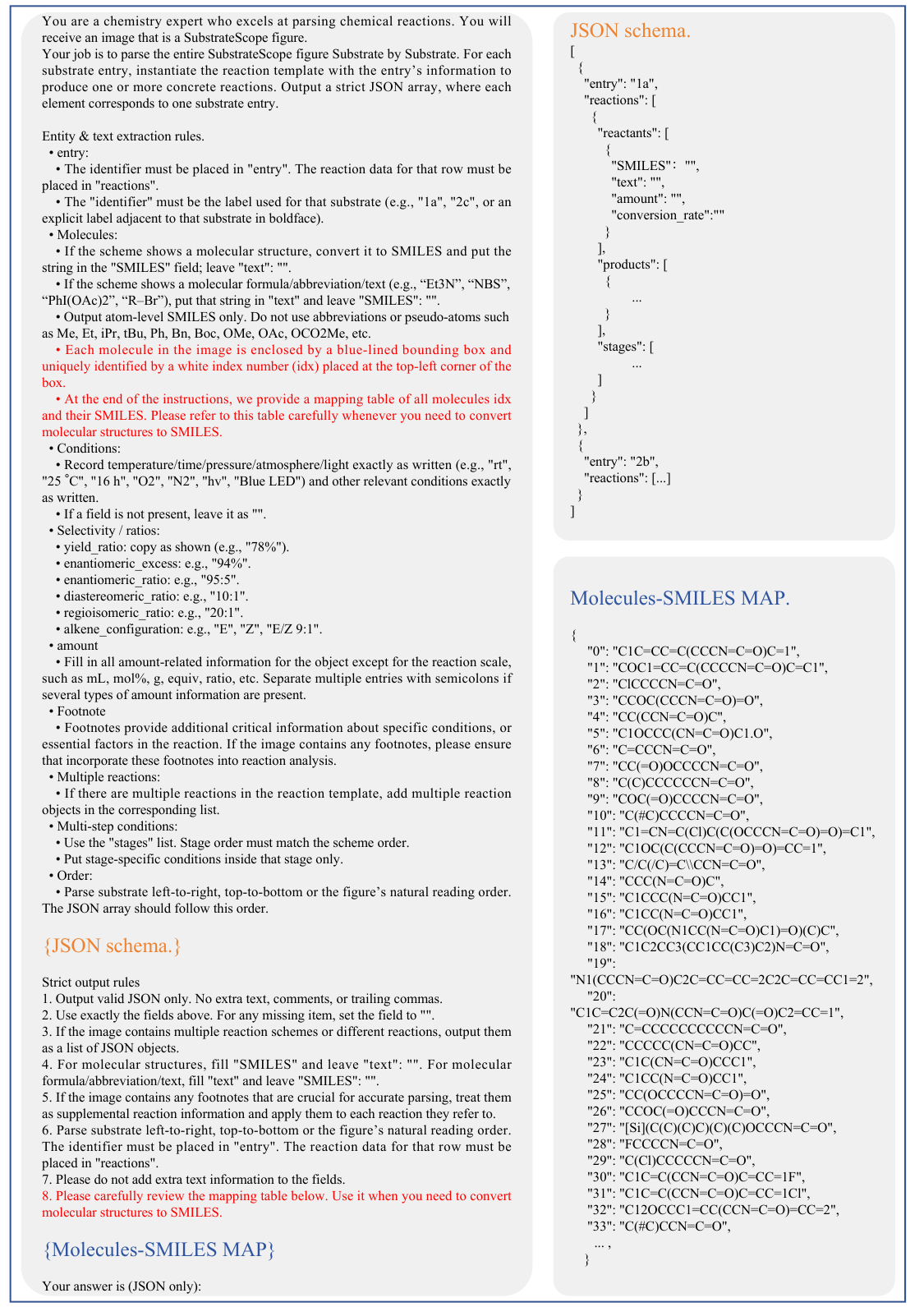}
    \caption{\textsf{Oracle-MolRec-BBOX-IDX inference} prompt of \textit{Substrate Scope}}
    \label{fig:Prompt_scope_idx}
\end{figure*}

\begin{figure*}[!t]
    \centering
    \includegraphics[width=0.9\linewidth]{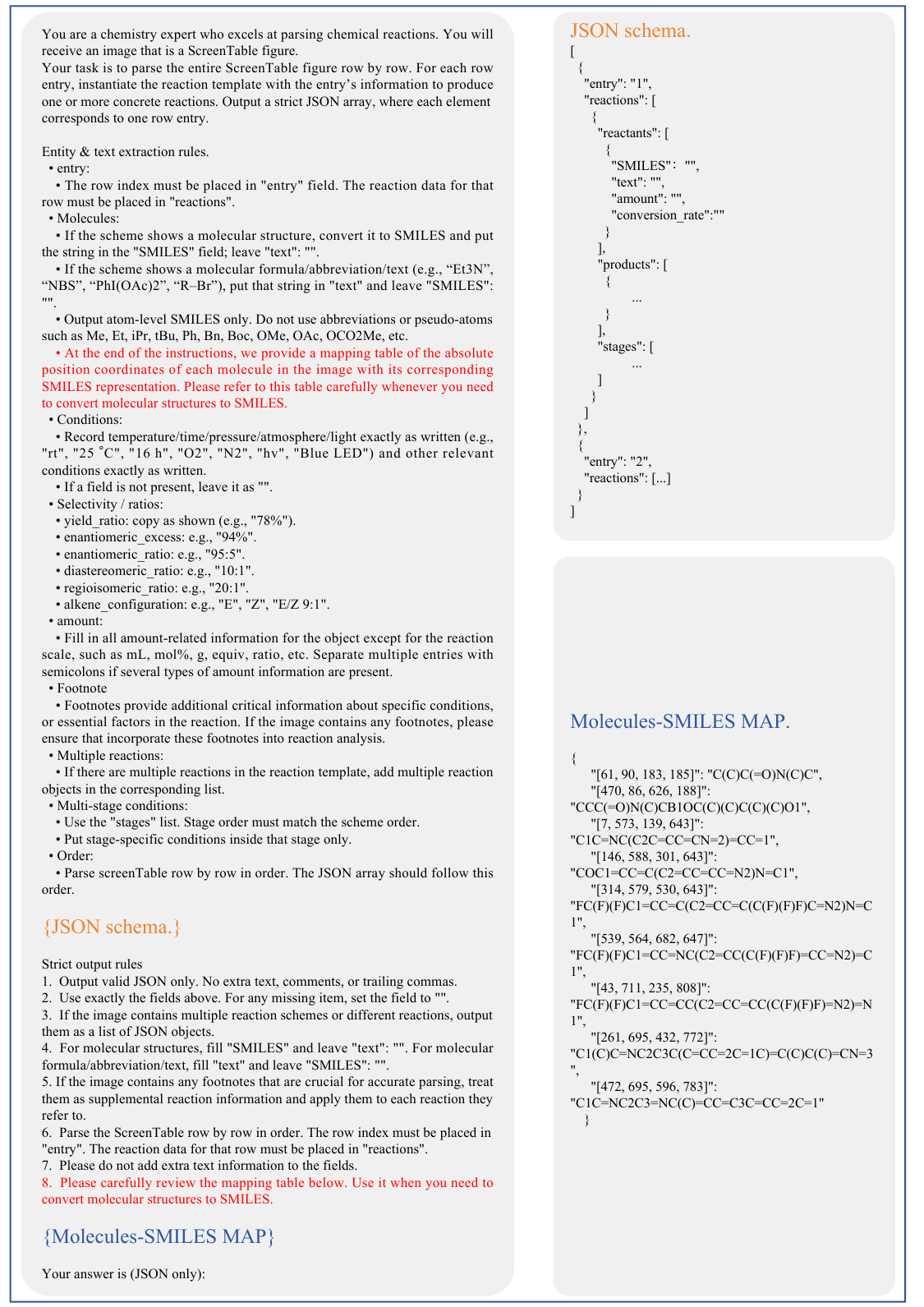}
    \caption{\textsf{Oracle-MolRec-BBOX-CORD inference} prompt of \textit{Screen Table}}
    \label{fig:Prompt_table_bbox}
\end{figure*}

\begin{figure*}[!t]
    \centering
    \includegraphics[width=0.9\linewidth]{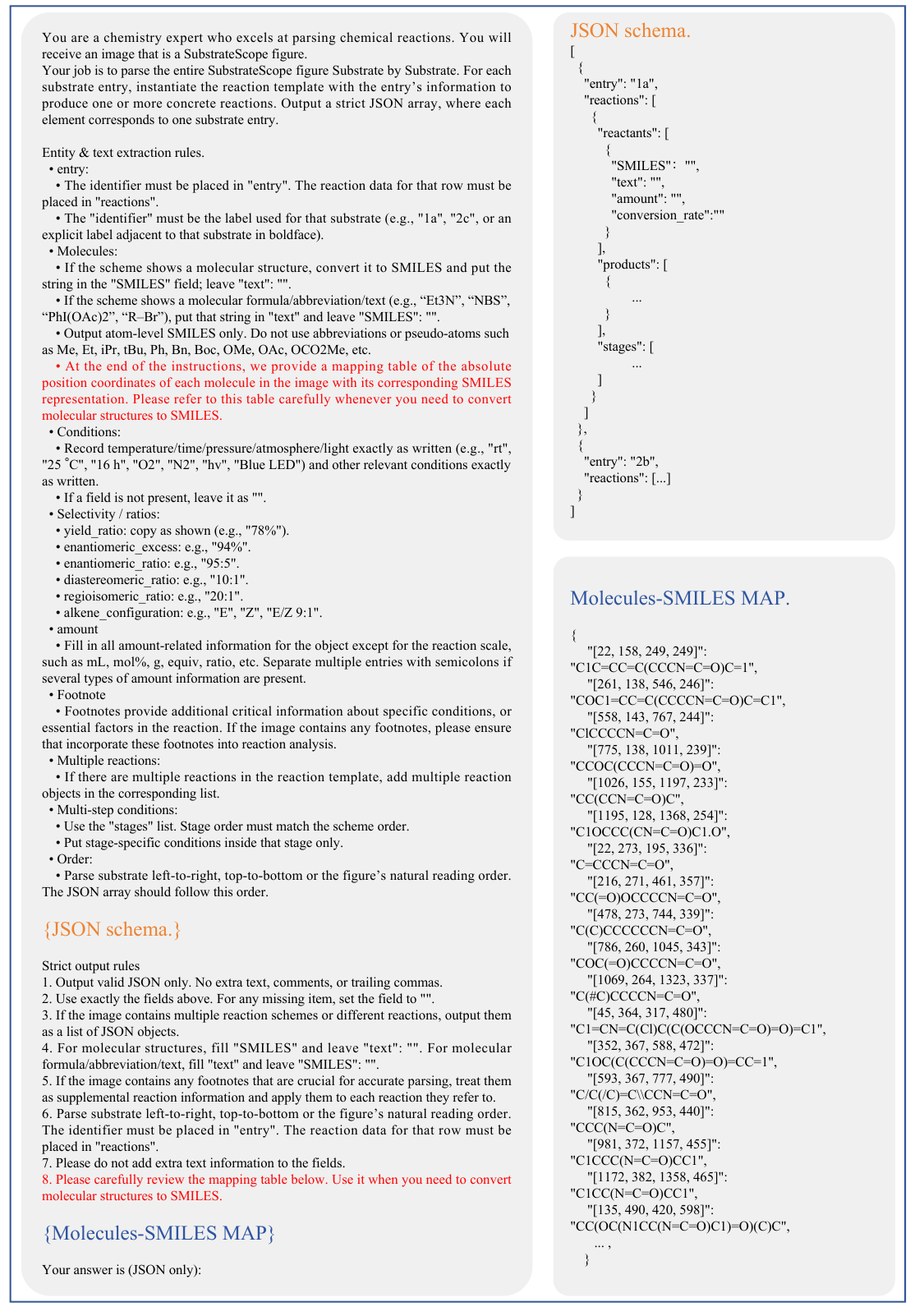}
    \caption{\textsf{Oracle-MolRec-BBOX-CORD inference} prompt of \textit{Substrate Scope}}
    \label{fig:Prompt_scope_bbox}
\end{figure*}

\begin{figure*}[!t]
    \centering
    \includegraphics[width=0.90\linewidth]{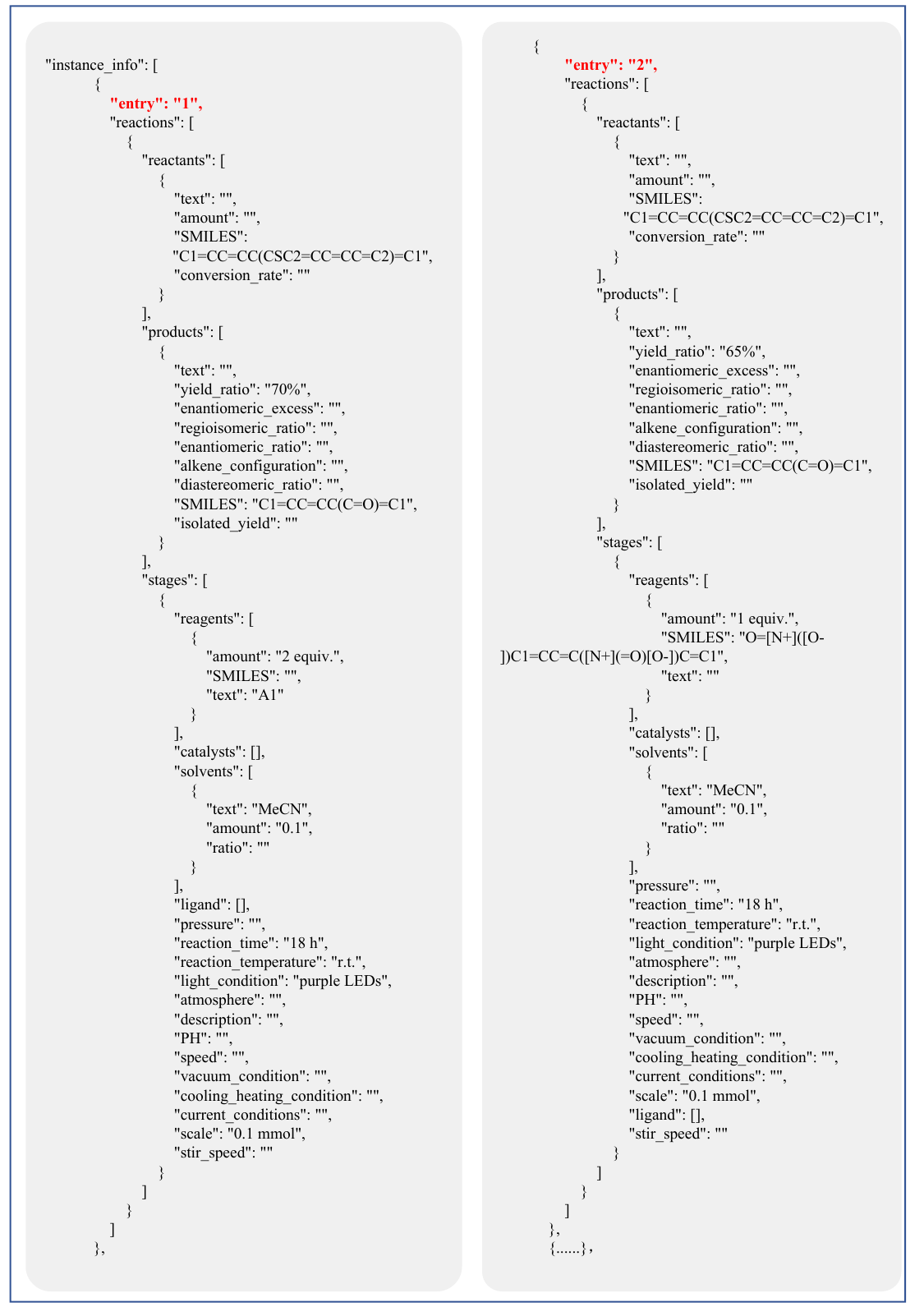}
     \caption{Annotation result of Screen Table.}
    \label{fig:GT_screentable1}
\end{figure*}

\begin{figure*}[!t]
    \centering
    \includegraphics[width=0.90\linewidth]{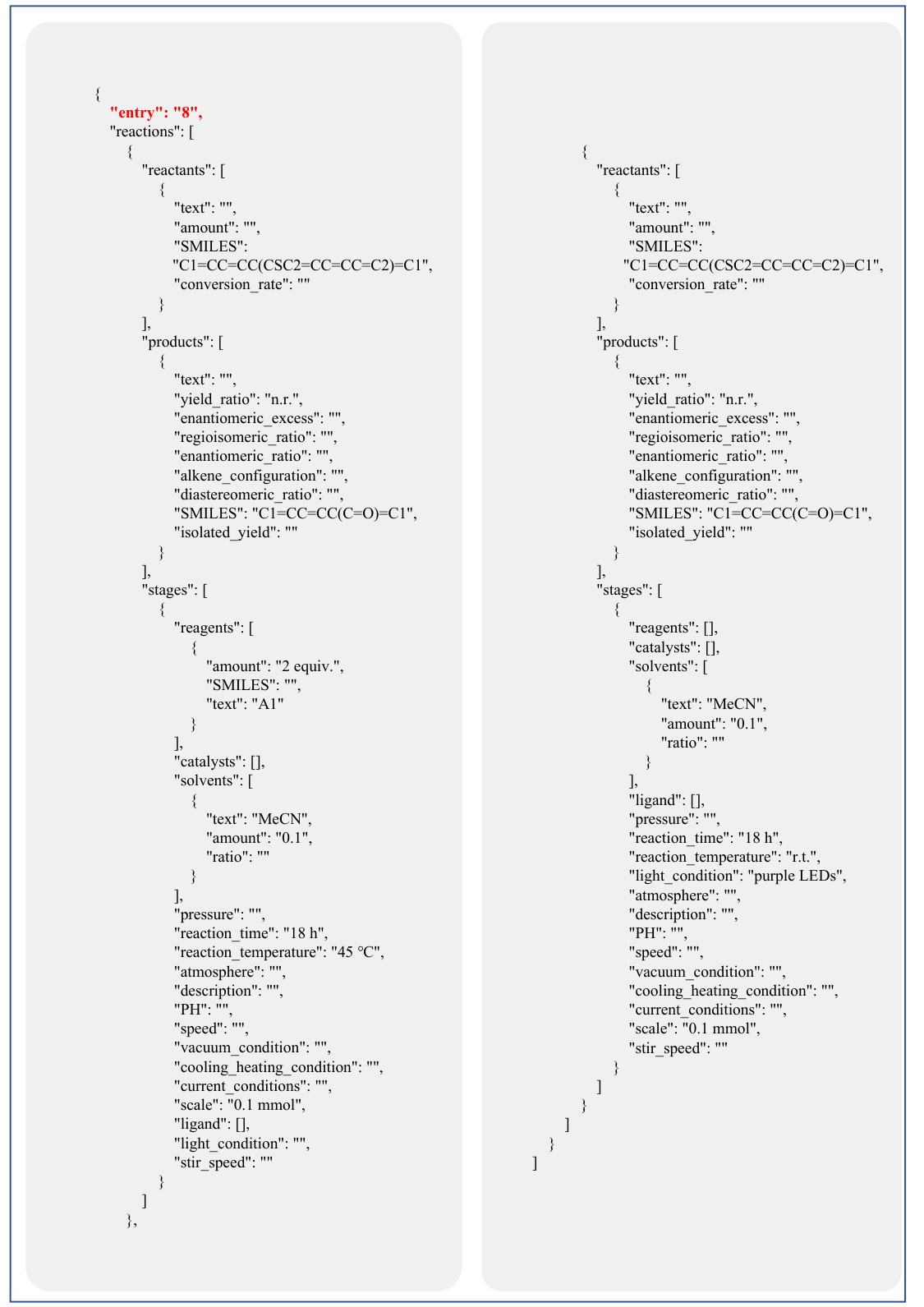}
    \caption{Annotation result of Screen Table.}
    \label{fig:GT_screentable2}
\end{figure*}

\begin{figure*}[!t]
    \centering
    \includegraphics[width=0.90\linewidth]{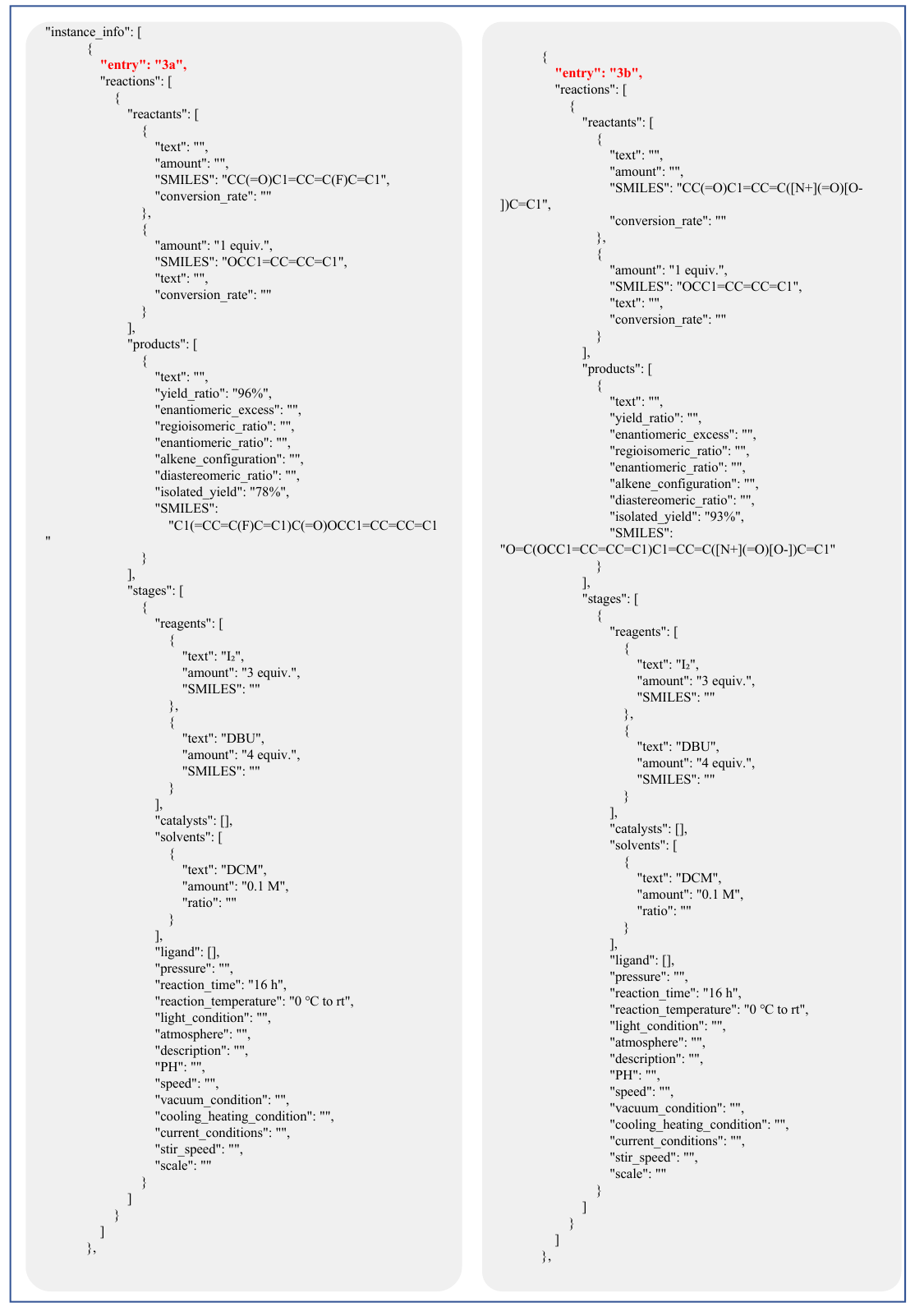}
     \caption{Annotation result of Substrate Scope.}
    \label{fig:GT_substrate1}
\end{figure*}

\begin{figure*}[!t]
    \centering
    \includegraphics[width=0.90\linewidth]{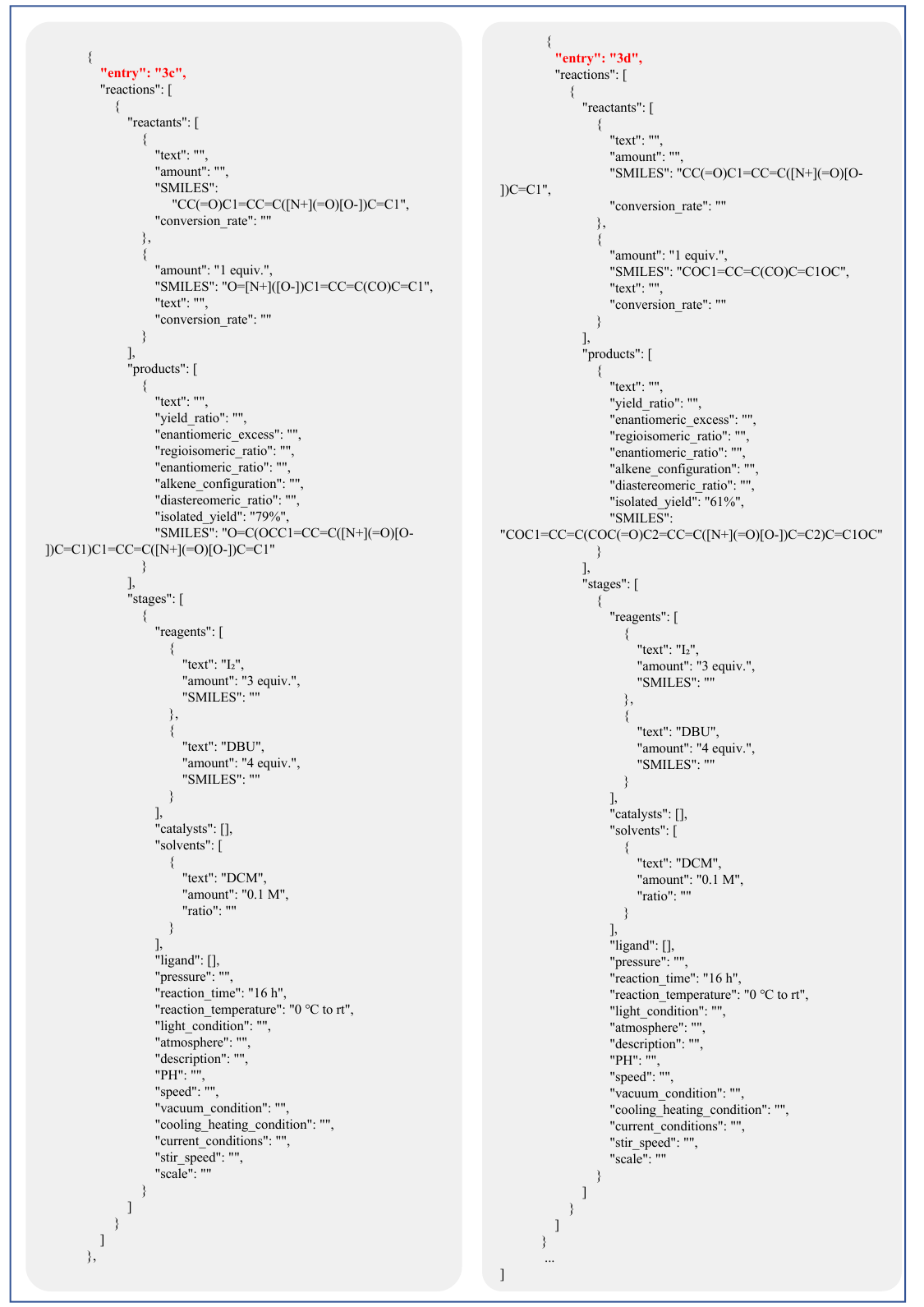}
    \caption{Annotation result of Substrate Scope.}
    \label{fig:GT_substrate2}
\end{figure*}

This subsection presents the detailed experimental results of RxnExBench. Table \ref{tab:Appendix_G_recall} shows the Recall scores of VLMs on RxnExBench, while Table \ref{tab:Appendix_G_precision} presents the Precision scores of VLMs on RxnExBench. The results are displayed for the full dataset and three difficulty levels under three evaluation criteria.
Figure \ref{fig:Appendix_G_all_venn_diagrams} presents the error type distributions of all models.

As described in Section~\ref{subsec:ablation_mol_rec}, the \textsf{Oracle-MolRec} protocol comprises three variants—\textsf{Oracle-MolRec-IDT}, \textsf{Oracle-MolRec-BBOX-IDX}, and \textsf{Oracle-MolRec-BBOX-CORD}—differing in how each molecule is referenced in the VLM prompt. An example input image with the bounding-box visual prompts used in \textsf{Oracle-MolRec-BBOX-IDX} is shown in Figure~\ref{fig:Prompt_vp_image}. The Recall, Precision, and F1 score results in Tables~\ref{tab:subset_recall},~\ref{tab:subset_precision}, and~\ref{tab:subset_f1} show that all three \textsf{Oracle-MolRec} variants significantly outperform \textsf{standard inference}, demonstrating the substantial value of accurate molecular structural information for reaction extraction. This also indicates that enhancing the parsing capability of molecular structures (SMILES) is crucial for VLMs in chemical reaction extraction.

%------------------------------------prompt------------------------------------%

%------------------------------------prompt------------------------------------%

\end{document}